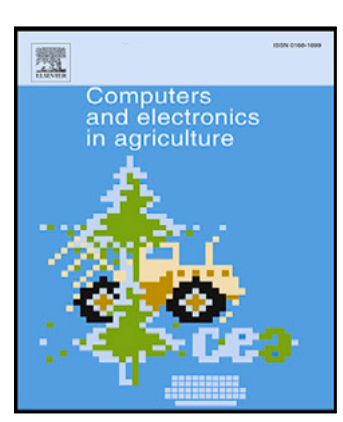

Original papers

# CottonSim: A vision-guided autonomous robotic system for cotton harvesting in Gazebo simulation

Thevathayarajh Thayananthan [a,b], Xin Zhang [a,b,*], Yanbo Huang [c], Jingdao Chen [d], Nuwan K. Wijewardane [b], Vitor S. Martins [b], Gary D. Chesser [b], Christopher T. Goodin [e]

[a] School of Environmental, Civil, Agricultural and Mechanical Engineering, University of Georgia, Athens, 30602, GA, USA
[b] Department of Agricultural and Biological Engineering, Mississippi State University, Mississippi State, 39762, MS, USA
[c] USDA-ARS, Genetics and Sustainable Agriculture Research Unit, Mississippi State, 39762, MS, USA
[d] Department of Computer Science and Engineering, Mississippi State University, Mississippi State, 39762, MS, USA
[e] Center for Advanced Vehicular Systems, Mississippi State University, Starkville, 39759, MS, USA

A B S T R A C T

Cotton is a major cash crop in the United States, with the country being a leading global producer and exporter. Nearly all U.S. cotton is grown in the Cotton Belt, spanning 17 states in the southern region. Harvesting remains a critical yet challenging stage, impacted by the use of costly, environmentally harmful defoliants and heavy, expensive cotton pickers. These factors contribute to yield loss, reduced fiber quality, and soil compaction, which collectively threaten long-term sustainability. To address these issues, this study proposes a lightweight, small-scale, vision-guided autonomous robotic cotton picker as an alternative. An autonomous system, built on Clearpath's Husky platform and integrated with the Cotton-Eye perception system, was developed and tested in the Gazebo simulation environment. A virtual cotton field was designed to facilitate autonomous navigation testing. The navigation system used Global Positioning System (GPS) and map-based guidance, assisted by an RGB-depth camera and a YOLOv8n-seg instance segmentation model. The model achieved a mean Average Precision ($mAP$) of 85.2%, recall of 88.9%, and precision of 93.0%. The GPS-based approach reached a 100% completion rate ($CR$) within a (5e-6)° threshold, while the map-based method achieved a 96.7% $CR$ within a 0.25 m threshold. The developed Robot Operating System (ROS) packages enable robust simulation of autonomous cotton picking, offering a scalable baseline for future agricultural robotics. CottonSim code and datasets are publicly available on GitHub: https://github.com/imtheva/CottonSim.

## 1. Introduction

The global population is projected to reach approximately 10 billion by 2050, according to a United Nations report (Population, 2024). The increase in the human population has led to higher demands for food and fiber. Modern agriculture faces significant challenges, such as the labor shortage (Barnes et al., 2021) and soil erosion (Montgomery, 2007), which pose threats to both agricultural security and sustainability. These challenges aggregate from various factors over time, including COVID-19 pandemic, migration, urbanization, and population aging (Rose et al., 2021). It is estimated that 70% of the population may live in urban areas by 2050, leading to a potentially even more severe labor shortage in agriculture (Phasinam et al., 2022). Digital agriculture is a farming strategy designed to provide farmers with precise and digitized data to make better informed decisions (Precision, 2023). Digital agriculture offers a promising mitigation for the emerging issues with advanced engineering technologies. In the past, farmers utilized the experience and judgment to make their own decisions and operated different types of machinery to manage their farms. Artificial intelligence (AI) and robotics are key components of digital agriculture, playing a crucial role in addressing labor shortages, economic constraints, and environmental challenges in modern farming. In recent years, autonomous robotic technologies have gained ample attention due to their accessibility and affordability, enabling key farming operations, such as weeding, sowing, planting, irrigating, and harvesting (Fasiolo et al., 2023; Oliveira et al., 2021). Ultimately, the goal is to have multiple lightweight, automated robots handle labor-intensive tasks, allowing farmers to upskill and focus on managing and maintaining robotic systems (Asseng and Asche, 2019). This shift underscores the urgent need for reliable, small-scale, and cost-effective autonomous solutions for various crops, including cotton.

* Corresponding author at: School of Environmental, Civil, Agricultural and Mechanical Engineering, University of Georgia, Athens, 30602, GA, USA.
*E-mail address:* xin.zhang2@uga.edu (X. Zhang).

Currently, almost all cotton in the U.S. is harvested with large and expensive mechanized harvesters that operate with either picking or stripping methods (Cotton Incorporated, 2025). These harvesters are highly efficient, requiring only one to two field workers to harvest approximately 809 ha (2000 acres) of cotton per year (Griffin and Barnes, 2017). However, conventional cotton harvesting have several limitations, such as degraded lint quality due to boll contamination. In contrast, the early-open bolls need to stay in the field (for about 40–60 days) while awaiting the late-open bolls (Barnes et al., 2021; Snipes and Baskin, 1994). Additionally, the weight of these harvesters is around 30 tonnes, which can contribute to soil compaction, negatively affecting soil health, reducing yield potential, and raising sustainability concerns (Gharakhani et al., 2023; Jamali et al., 2021; Majdoubi et al., 2024).

Beyond agronomic concerns, the financial burden of large-scale harvesters limits their accessibility for smaller farms. A modern six-row John Deere (Deere & Company, Moline, IL, USA) module cotton harvester, for instance, costs around $1 million (Deere, 2025; All-Machines, 2025). To justify this investment, a farmer would need to manage 500–600 ha (1200–1500 acres) of cotton production, assuming a production cost of $1985 per hectare per year, based on the USDA's 2022 cotton cost-of-production forecast (Wilde, 2022). The high costs of purchasing and upkeeping large-scale harvesters further add to the burden (Barnes et al., 2021; Gharakhani et al., 2023). Moreover, traditional cotton harvesters require the use of chemical defoliants before harvesting, which could have environmental and economic drawbacks.

The future vision for precision agriculture is to implement autonomous systems and sustainable techniques in cotton farming, potentially employing a swarm of intercommunicated small-scale (Sperati et al., 2011), low-cost, and lightweight robots working together to carry out field tasks such as cotton harvesting as soon as the cotton bolls crack open and become ready for harvest. These autonomous systems could enable timely cotton picking, eliminate the need for defoliant chemicals, and mitigate soil compaction issues. Equipped with self-navigation and path-planning capabilities, a swarm of robots could efficiently assist in cotton harvesting while reducing the environmental footprint.

To initiate development, a lightweight and autonomous cotton-picking system must first be designed and validated. Simulation studies are widely recognized as essential tools in robotics and automation research (Žlajpah, 2008). They allow for the design and evaluation of conceptual models and algorithms before physical implementation. Compared to direct field deployment, simulations offer several advantages, including reduced cost and risk, faster iteration cycles, improved scalability, performance tuning, and valuable educational opportunities (Choi et al., 2021; Sotiropoulos et al., 2017).

The use of robot simulation in agriculture has been widely explored for testing robot design, navigation control, field mapping, and crop monitoring. Among simulation environments, Gazebo, an open-source simulator integrated with the Robot Operating System (ROS), is one of the most commonly adopted platforms. It provides a comprehensive robotics simulation toolbox featuring a plugin-based architecture for defining sensors, actuators, and environments. Gazebo operates through the *gzserver* engine and incorporates realistic physics using the Open Dynamics Engine (ODE), making it highly suitable for simulating robotic systems in diverse agricultural scenarios.

A research group tested reactive row navigation in both indoor and outdoor simulation environments using the ROS-based Gazebo platform (Linz et al., 2014). Their robot model incorporated physical characteristics such as friction and inertia to reflect realistic motion. Real-farm data, such as odometry, point cloud information, and video frames, were collected from orchards and vineyards and used to fine-tune the navigation stack parameters through iterative testing in the simulated environment.

In greenhouse settings, another research team simulated an indoor farming environment to implement SLAM-based autonomous navigation using algorithms such as RTAB-Map (Real-Time Appearance-Based Mapping) and grid mapping (Khan et al., 2021). Their objective was to enable a fully autonomous robot to explore greenhouse rows and generate accurate 2D maps. Another study, focusing on greenhouse environments, was completed to design and develop an agricultural robot localization system in ROS-based Gazebo simulator with a Turtlebot3 platform (ROBOTIS Co. Ltd, Seoul, South Korea) employing the Gmapping algorithm-based SLAM (Ratul et al., 2021). To enhance realism, a virtual greenhouse was created using Blender (Blender Institute, Amsterdam, The Netherlands), allowing the system to be tested under spatial constraints similar to real-world environments.

In orchard applications, virtual apple and palm oil orchards in Gazebo were developed to evaluate 2D field mapping performance. Compared to Hector-SLAM, SLAM-Gmapping demonstrated superior accuracy and map consistency. Therefore, the proposed system integrated SLAM-Gmapping for localization and mapping, combined with tree and fruit detection, as well as volumetric mapping based on tree properties, to generate a comprehensive orchard map (Habibie et al., 2017). The volumetric SLAM-Gmapping approach, paired with a detection algorithm, showed strong performance, highlighting Gazebo's suitability for evaluating SLAM techniques in agricultural landscapes.

Simulation has further been applied in indoor yield estimation tasks as well. One study developed an aerial robotic system for sweet pepper yield mapping, integrating RGB-depth (RGB-D) data acquisition with a pretrained MobileNet Single Shot Detector (SSD) and the OPTICS clustering algorithm (Ivanovic et al., 2022). The simulation incorporated realistic trajectory control and vision-based perception, enabling effective fruit detection and counting in a structured greenhouse setup.

In cotton phenotyping applications, a ROS-based field robot was equipped with a 2D Light Detection And Ranging (LiDAR) sensor in a nodding configuration to generate 3D point cloud data (Iqbal et al., 2020). The robot navigated a virtual Gazebo field containing multiple crop plots by combining LiDAR-based local control with GPS-guided Pure Pursuit for inter-row navigation. This hybrid approach allowed effective phenotyping in occluded or misaligned crop rows and ensured accurate movement across multiple zones in the field.

A modular ground robot for cotton phenotyping, MARS-PhenoBot, has also been developed in Gazebo for testing autonomous navigation in uneven terrain (Li et al., 2022). The platform utilized crop row structure to guide movement, avoiding the need for GNSS or pre-mapped localization. The system used YOLOv4-tiny and DeepSORT for object detection and tracking, and successfully generated 2D and 3D maps of crop rows. Although odometry drift limited the accuracy of long-term 3D mapping, the methodology proved scalable for practical deployment.

However, there remains a notable gap in simulation-based methodologies tailored specifically to autonomous navigation and cotton harvesting within Gazebo. In this study, we present CottonSim, the first comprehensive Gazebo-based simulation framework focused on vision-guided autonomous navigation and picking in cotton farming. The primary objective is to develop and validate a modular robotic system for selective cotton harvesting using both GPS-based and map-based navigation strategies within a virtual cotton field. More specifically, this study aims to: (1) design and develop a lightweight autonomous cotton-picking robot and a virtual cotton farm in the ROS-based Gazebo simulator, (2) integrate the Cotton-Eye perception system for real-time detection and alignment, and (3) evaluate the visual-guided navigation strategies, along with the map and GPS-based autonomous navigation approaches, of the robotic system in the virtual cotton farm.

To advance realistic field navigation, we constructed a detailed simulation environment to test the robot's capability to traverse cotton rows and perform precise movements during harvesting. This work represents the first simulation-based investigation of vision-guided autonomous cotton harvesting in the ROS-Gazebo platform. By incorporating deep learning-enabled instance segmentation for path correction and alignment, the system supports enhanced autonomous behavior under field-like conditions. Performance was analyzed using both GPS-based and map-based navigation techniques. This virtual testbed provides an essential platform for refining system behavior and ensuring robustness before physical deployment.

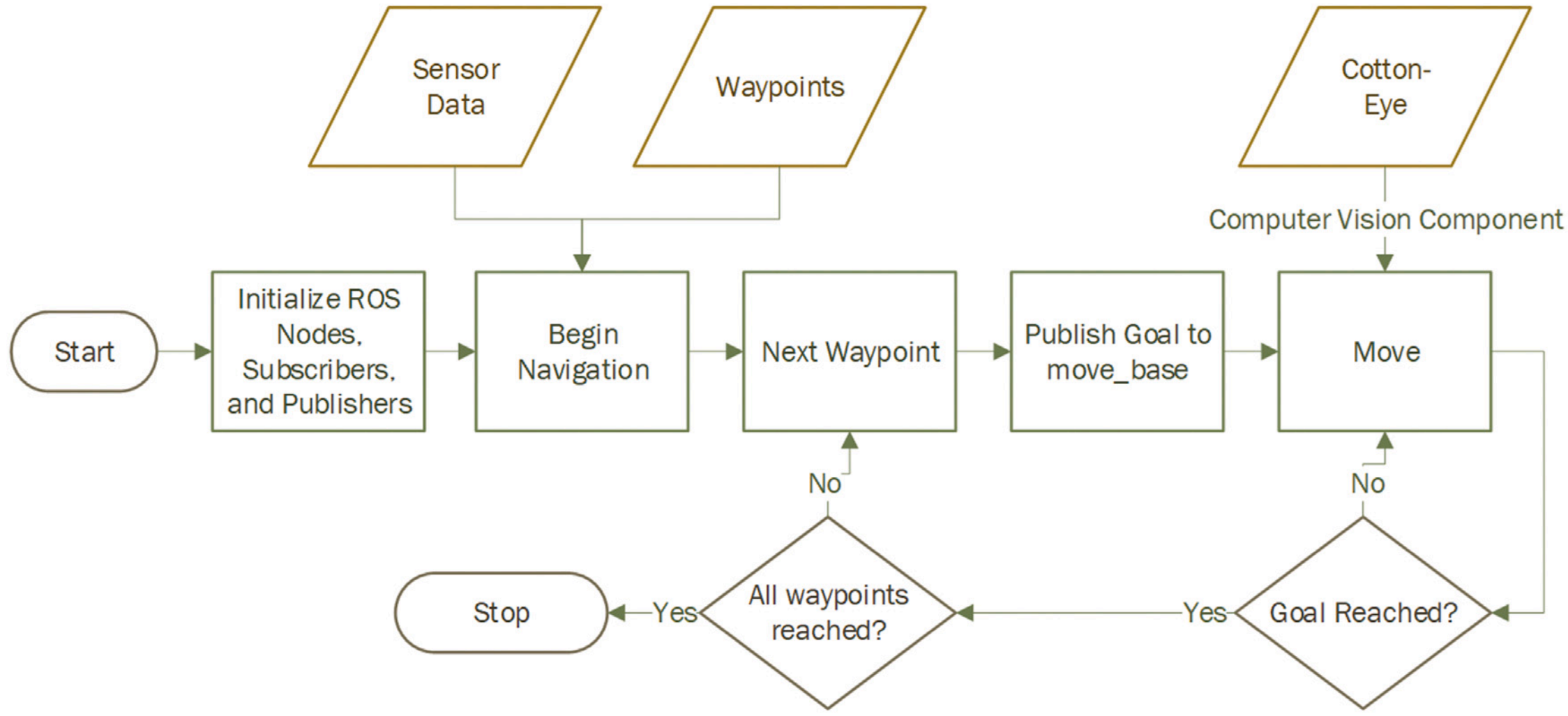

**Fig. 1.** Overall workflow diagram of *CottonSim*.

The proposed methodology and implementation details are described in the following subsections: Approach Overview, Simulation Setup with ROS and Gazebo, Virtual Cotton Farm Generation, Cotton Picker Robot Development, Cotton-Eye Perception System, Robot Navigation Control, and Evaluation Metrics.

## 2. Methodology

### 2.1. Approach overview

Fig. 1 illustrates the overall workflow of the *CottonSim* for the autonomous navigation of the robotic cotton picker. The simulation starts by initializing and setting up the ROS nodes, subscribers, and publishers. Sensory data are continuously processed to update the navigation stack, ensuring obstacle avoidance and path correction.

Two different types of waypoints are handled by two navigation approaches: map-based and GPS-based navigation. Map-based waypoints are directly published to *'move_base'*, which is a ROS package, to facilitate the navigation. In contrast, GPS-based waypoints, provided in latitude and longitude coordinates, are first converted to Universal Transverse Mercator (UTM) coordinates. These UTM coordinates are then transformed to the map frame using the Transform2 library. Similar to map-based approach, the transformed coordinates are then used to generate the navigation goals, which are published to *move_base*.

Once the navigation goals are published in ROS, the robot begins moving toward the designated waypoints. The Cotton-Eye perception module (Cotton-Eye Perception System section) assists with the cotton-picking robot, ensuring the robot remains centered within the crop rows. Upon reaching a waypoint, the system sequentially publishes the next waypoint until the final destination is reached. Once the last waypoint is reached, the robot stops.

### 2.2. Simulation setup with ROS and Gazebo

The development of the simulation environment was implemented using the Robot Operating System (ROS) and the Gazebo robotic simulator on the Ubuntu 20.04.5 LTS (Focal Fossa). ROS is an open-source framework that provides software developers with libraries and tools to create robotic applications. ROS Noetic (Noetic Ninjemys), one of the ROS distributions, was utilized in this research for simulation development. ROS distributions contain numerous ROS packages that facilitate developers to work with a stable code base. ROS Noetic is the 13th released ROS distribution and the final release for ROS 1, targeting the Ubuntu 20.04 release. Additionally, the Gazebo 11.0.0 version was used to perform all simulation tasks using the developed ROS scripts and meshes. RViz was also utilized along with the Gazebo robotic simulator to visualize sensor data and monitor the movement of the cotton-picking robot. RViz is a graphical interface in ROS that enables the visualization of various types of data using plugins for multiple available topics.

*CottonSim* code and data are released to the research community via GitHub (https://github.com/imtheva/CottonSim).

### 2.3. Virtual cotton farm generation

The development of the virtual cotton farm was initiated by acquiring a 3D model of cotton plant (Treant, 2023) from the open-source 3D model marketplace CGTrader. Among all available cotton plant models, the plant (Fig. 2) with fully-developed cotton bolls before defoliation (i.e., cotton cut out stage) was selected for the virtual farm development. This is the cotton growth stage where most of the leaves and cotton bolls are present before the defoliation. This stage is particularly challenging as the leafy canopy may negatively affect the navigation capability of the robot. Therefore, it was selected for testing purposes in this study. Our proposed technique also can be adapted for other growth stages of cotton, including after the defoliation. The height and width of each individual cotton plant are 1.00 m (39.37 in) and 0.70 m (27.56 in), respectively, as depicted in Fig. 2. This cotton model was then imported into the Blender 3.5 to develop the whole virtual cotton farm.

Since the virtual cotton farm needs to be developed in a way that ensures the cotton-picking robot, equipped with a 3D LiDAR, to traverse and navigate without encountering any obstacles. Cotton row spacing is an important factor to consider. If the row spacing is too narrow, the robot may not be able to enter the cotton farm because the LiDAR would detect the extended canopies as obstacles and stop the robot. Fig. 3a showcases ten different row spacings, ranging from 1.00 m (39.37 in) to 1.90 m (74.80 in) with an increment of 0.10 m (3.94 in), that were tested to determine the minimum row spacing required for the robot to traverse. The center of each row spacing was used as the reference point. The farm configuration was constructed in the Blender 3.5 (Fig. 2), and the simulation was performed using the Gazebo.

With the given widths of the cotton plant and the cotton-picking robot, 0.67 m (26.38 in), which is described in the next section Cotton

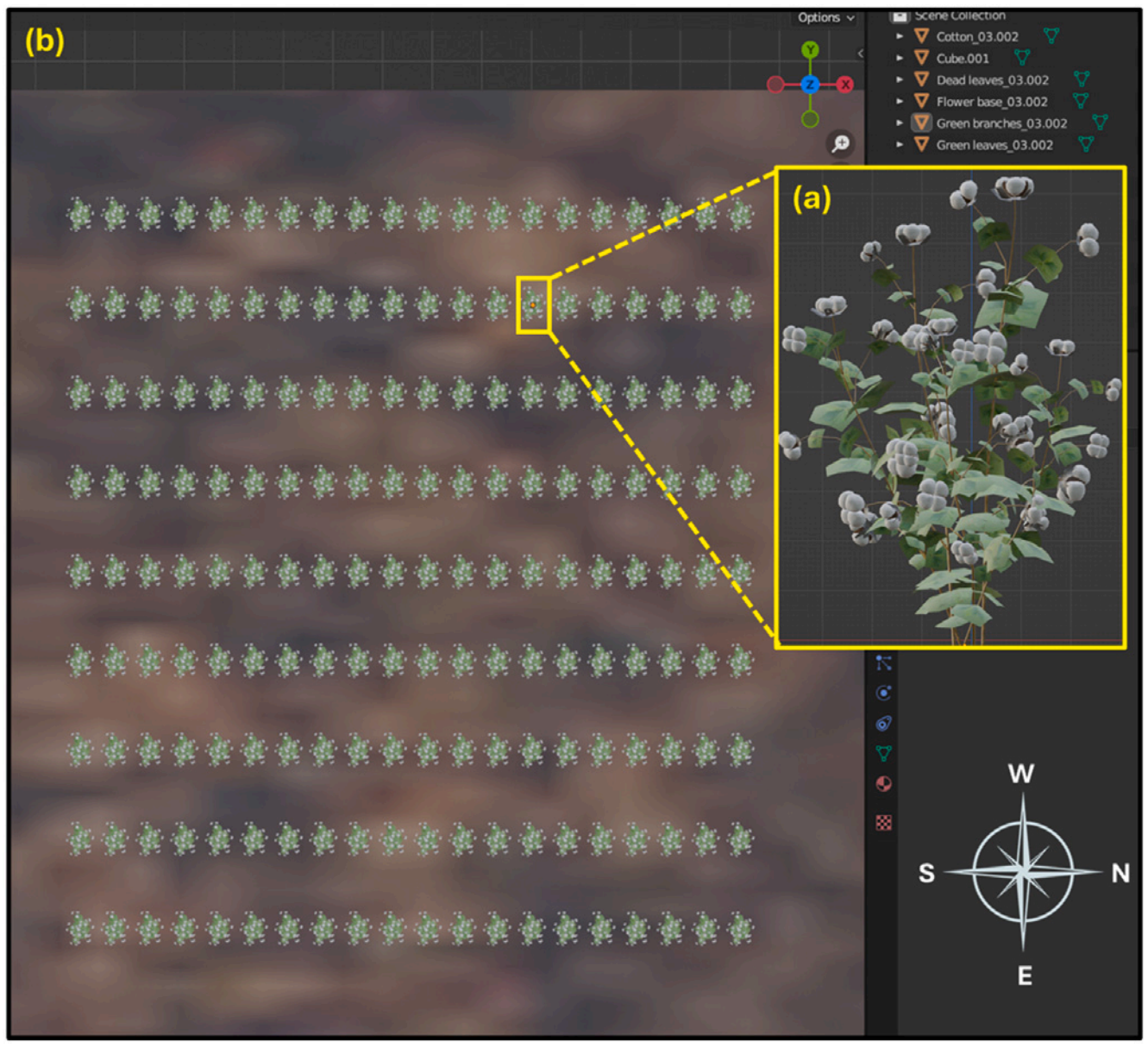

**Fig. 2.** Development of the virtual cotton farm: (a) an acquired 3D model of cotton plant before defoliation (Treant, 2023) and (b) the overall cotton farm configuration in Blender 3.5.

Picker Robot Development, the test results suggested that the robot could smoothly traverse the farm at a row spacing of 1.80 m (70.87 in). Therefore, a virtual cotton farm was developed with this row spacing, as depicted in Fig. 2a. To be computationally efficient, the virtual farm comprised of only one plot, including nine rows of cotton (along the north–south (N–S) direction) with 20 plants per row using the suggested row spacing. The plant spacing within each row was set to 0.70 m (27.60 in) (Fig. 3b). The overall setup of the virtual cotton farm, where the total plant area is 222.76 $m^2$ and the total farm area is 1233.97 $m^2$, is illustrated in Fig. 2a.

Once the virtual cotton farm's geometry and row spacing were properly defined, individual plant meshes were then converted to the Unified Robotic Description Format (URDF) files to store physical properties and other object metadata. The meshes of cotton plants and the world, including *cotton_world.dae, cotton_branch.dae, cotton_leaves.dae, cotton_dead_leaves.dae, cotton_flower_base.dae, and cotton.dae*, were integrated as visual elements (where a .dae file, or Digital Asset Exchange file, is a standard file format for exchanging mesh data). Among all visual elements, *cotton_branch.dae, cotton_dead_leaves.dae, and cotton_leaves.dae* were defined as collision elements, allowing these components to interact with the robot as obstacles in the simulation. All plant meshes used were set with a specified origin and orientation, ensuring their alignment within the virtual cotton farm. The collision properties, including the primary ($\mu_1$ = 100) and secondary ($\mu_2$ = 50) friction coefficients, were defined to simulate realistic physical interactions between the robot and soil by tuning the resistance of vehicle movement.

Our comprehensive modeling of the cotton farm provides a reliable prerequisite for accurate simulations in the Gazebo, while considering the physical constraints and robot-plant interactions for autonomous navigation of a cotton-picking robot.

### 2.4. Cotton picker robot development

The cotton picker robot was developed on top of a rugged, outdoor-ready uncrewed ground vehicle (UGV), Husky A200 (Clearpath Robotics, Ontario, Canada; parent company: Rockwell Automation, Inc., Milwaukee, WI, USA) (Husky, 2024). The Husky UGV features a compact yet robust design, measuring 990 × 670 × 390 mm (39.00 × 26.40 × 14.6 in) in length, width, and height, with a maximum of 75 kg (165 lbs) payload capacity and four high-torque drive motors suitable for rough terrain. It is powered by a 24 V 20 Ah sealed lead acid battery, providing approximately 3–5 h of continuous operation depending on load and usage. Additionally, the platform supports various sensor integrations, including LiDAR, stereo cameras, and GPS modules, facilitating autonomous navigation and real-time perception for agricultural tasks.

To enable the visual-guided robotic cotton-picking system navigating autonomously in our developed virtual cotton farm, the Husky's URDF file was obtained from Clearpath Robotics' GitHub repository (GitHub, 2023). Before importing it into the Gazebo robotic simulator, the system integration was completed based on the physical robotic system in our laboratory. First, the Cotton-Eye perception system was integrated, where three RealSense RGB-D cameras (Intel Corporation, Santa Clara, CA, USA) were used, including a front-facing primary camera for visual-guided navigation and two left-facing (secondary) and right-facing (tertiary) cameras for assisting cotton-picking, the later of which is a planned feature for future implementation. Second, a Velodyne VLP-16 LiDAR (Velodyne Lidar, San Jose, CA, USA) was integrated with the system for 2D field mapping (with point cloud data) and assistive navigation in the virtual cotton farm. Third, the URDF of a UR5e robotic arm (Universal Robots, Odense, Denmark) (Husky Manipulation, 2023) and its control box were integrated with the Husky's URDF file (Fig. 4), where the robotic picking mechanism is planned to be explored in the future (Future Work section). The overall configuration of the cotton-picking robot is shown in Fig. 4. Last, a vision-RTK (RTK: Real-Time Kinematic) (Fixposition AG, Schlieren, Switzerland) was also added to realistically mimic our physical cotton-picking robot. Vision-RTK combines RTK-GNSS technology with computer vision and inertial sensor to deliver accurate global positioning and pose estimation even in GNSS-compromised environments such as under tree canopies or near tall crop rows. By fusing camera images and inertial measurements with GNSS data using advanced sensor fusion algorithms, the vision-RTK ensures reliable and high-precision localization throughout the robot's navigation path in the field, thus addressing common edge cases where traditional GNSS solutions fail. The vision-RTK is installed on the physical robot to properly localize the system in the farm environments (Meier, 2023).

The relative positions and poses of the robot components, including sensors and decorations, with reference to the center of Clearpath Husky's top plate frame (Fig. 4) are given in Table 1. The top plate frame was located at X = 0.081 m (3.197 in) and Z = 0.245 m (9.646 in) from the robot's base plate coordinate frame.

The Cotton-Eye perception system (Cotton-Eye Perception System section) consists of three RealSense RGB-D cameras. The front-facing, primary camera was placed at the front of the cotton picker where its relative position and pose from the center of the top plate (reference frame) are given in Table 1. The camera was pointing downwards to better perceive the farm environment within its field-of-view (FOV) of 69° × 42° (horizontal × vertical), including the cotton plants, ground, and sky. This camera was mainly used to facilitate the movement of the robot by continuously center-aligning the cotton picker inside the cotton rows along with the generated waypoints using either map-based or GPS-based navigation approach (Robot Navigation Control section). The primary camera model was configured to publish data under the ROS topic namespace *'realsense_p'*. Table A.4 summarizes a few main ROS topic namespaces and their associated topics that were published during the robot's autonomous navigation in Gazebo. Additionally, other relevant topics, including heading information, were managed by the IMU and the Clearpath Husky's control and navigation stack (GitHub, 2023).

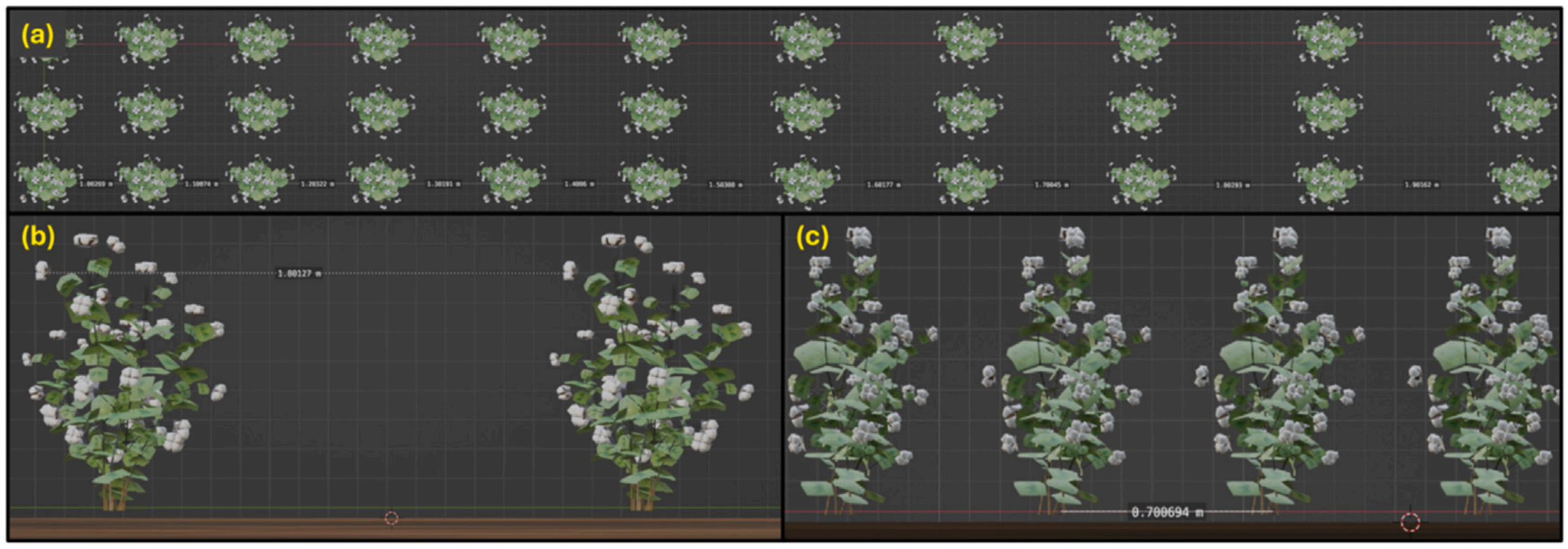

**Fig. 3.** Development of the virtual cotton farm: (a) testing ten different row spacing of cotton plants and final selections of (b) the row spacing of 1.80 m (70.87 in) and (c) the plant spacing of 0.70 m (27.60 in) within each row in the virtual cotton farm.

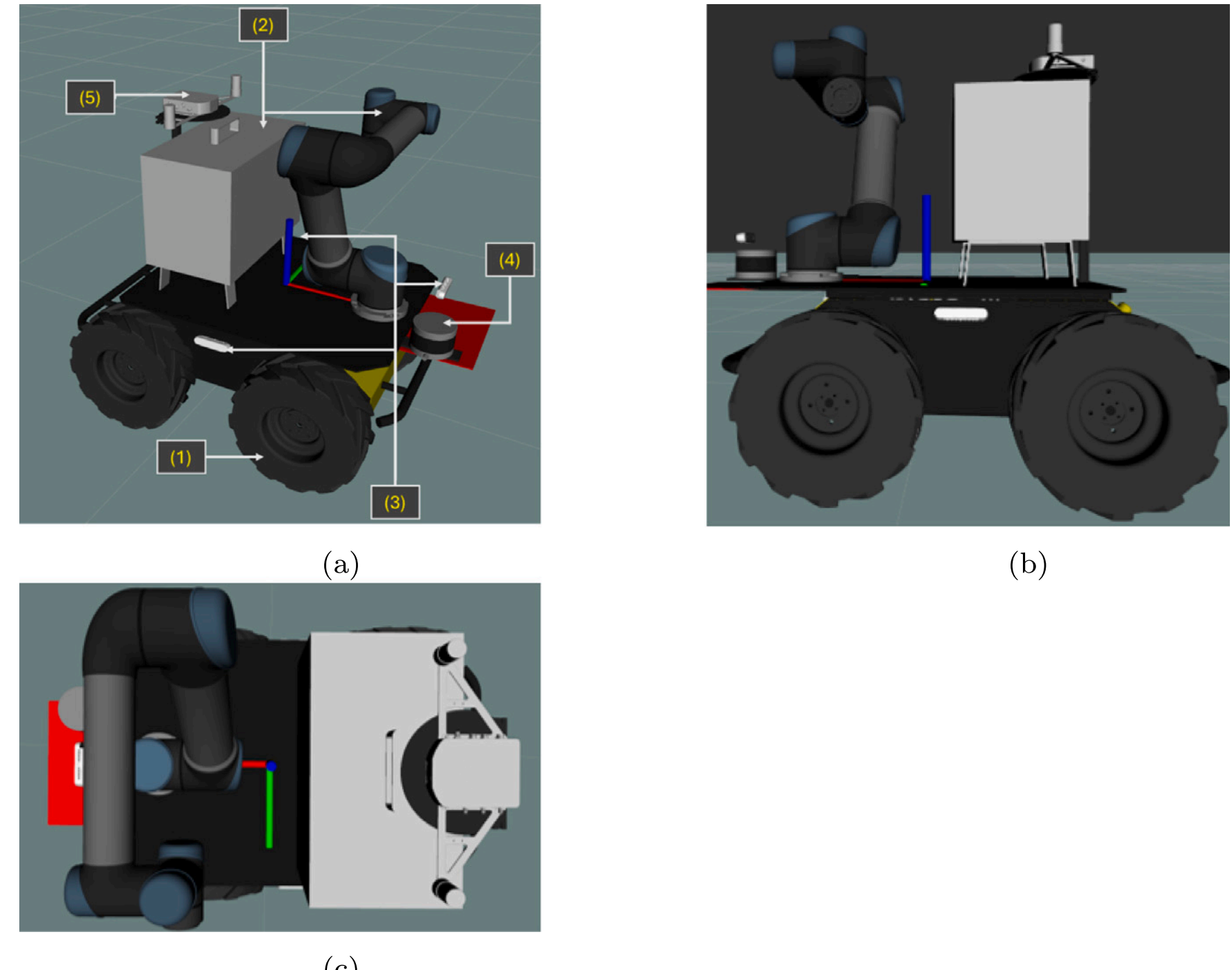

(a) (b)

(c)

**Fig. 4.** The cotton-picking robot represented using Unified Robotic Description Format (URDF): (a) system overview ((1) a Clearpath Robotics' Husky platform; (2) a Universal Robots' UR5e robotic arm and control box; (3) the Cotton-Eye perception system with three RealSense RGB-D cameras; (4) a Velodyne 3D LiDAR; and (5) a Fixposition's vision-RTK), (b) side view, and (c) top view.
The XYZ coordinate frame depicted in each sub-figure represents the Clearpath Husky's top plate coordinate frame, where other components are referenced to (Table 1). Its orientation is denoted using X, Y, and Z, respectively.

**Table 1**
The relative positions and poses of the robot components with reference to the center of Clearpath Husky's top plate frame (Fig. 4).

| Robot component | X (m) | Y (m) | Z (m) | Roll (°) | Pitch (°) | Yaw (°) |
|---|---|---|---|---|---|---|
| Primary camera | 0.415 | 0.000 | 0.095 | 0 | 5 | 0 |
| Secondary camera | −0.070 | 0.300 | −0.050 | 0 | −15 | 90 |
| Tertiary camera | −0.070 | −0.300 | −0.050 | 0 | −15 | −90 |
| LiDAR | 0.430 | −0.125 | 0.000 | 0 | 0 | 90 |
| IMU | 0.109 | 0.000 | −0.096 | 0 | −90 | 180 |
| GPS | −0.354 | 0.190 | 0.513 | 0 | 0 | 0 |
| UR5e arm (Base) | 0.264 | 0.000 | 0.006 | 0 | 0 | −90 |
| Base plate (Center) | −0.081 | 0.000 | −0.245 | 0 | 0 | 0 |

The side-facing, secondary and tertiary cameras were symmetrically positioned on the left and right sides of the robot (Fig. 4(a)). The secondary and tertiary cameras were placed with a distance of 0.600 m (23.622 in) opposite to each other, and their relative positions and poses to the top plate coordinate frame (reference) were listed in Table 1. The configuration of these two cameras is crucial to cover the entire cotton canopy within their FOV for subsequent cotton boll detection and cotton-picking on both sides. The models of the secondary and tertiary cameras were configured to publish data via the rostopic namespaces of *'realsense_secondary'* and *'realsense_tertiary'*, respectively.

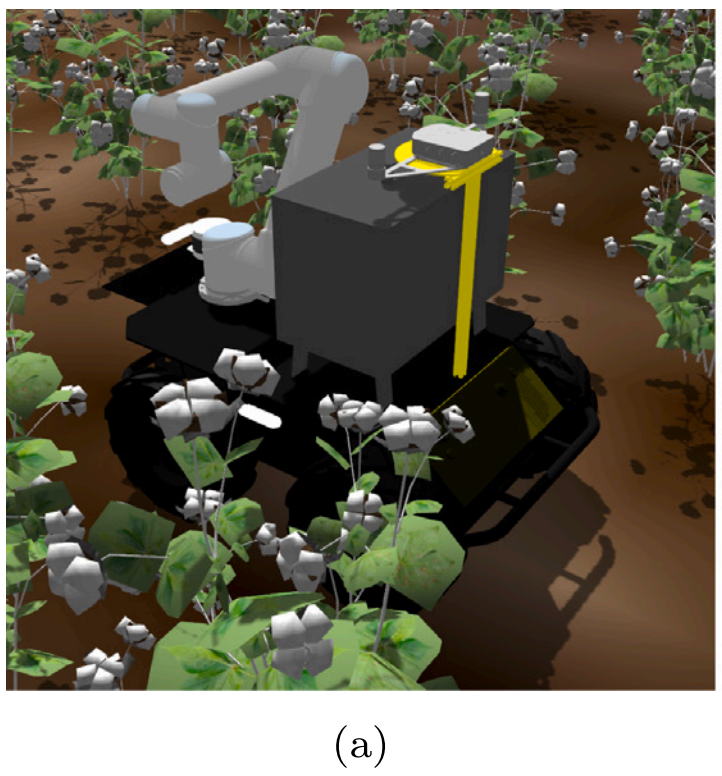

(a)

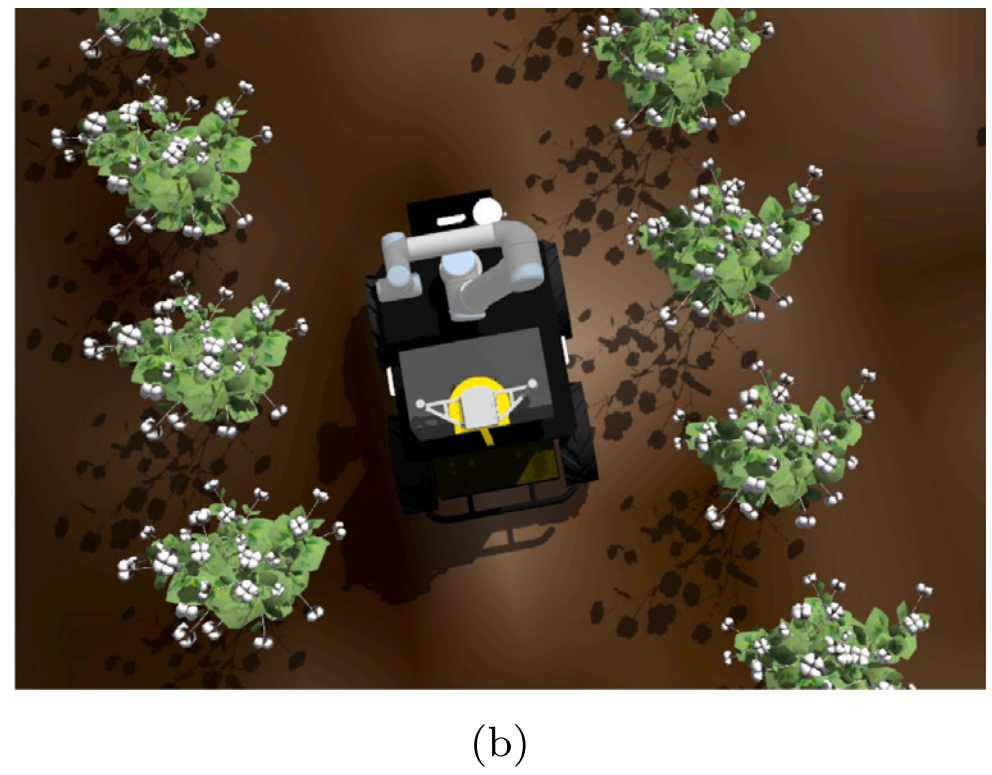

(b)

**Fig. 5.** Demonstrations of the integrated cotton-picking robot performing autonomous navigation in the Gazebo virtual cotton farm: (a) side-view and (b) top-view.

A GPS sensor was installed at the rear of the Husky to continuously obtain the correct geo-locations in latitudes and longitudes, while an inertial measurement unit (IMU) sensor was installed inside the robot to facilitate navigation by measuring orientation, velocity, and acceleration. The positions and poses of these two sensors relative to the top plate coordinate frame of the Husky are given in Table 1.

The Velodyne 3D LiDAR, mounted at the front of the Husky (Fig. 4), was also employed to enable the localization and obstacle avoidance along the robot's traversal path. The point cloud data (PCD) generated by the 3D LiDAR played an essential role in supporting costmap-based localization and path planning for the robot. A costmap is a grid-based approach that assigns cost values to regions based on obstacles and free space, enabling the robot can navigate efficiently. A customized ROS launch script was developed to apply a PassThrough filter, limiting the PCD from the 3D LiDAR within the $Y$-axis range (i.e., −10 to 0 m (−393.701 to 0 in)) and the $Z$-axis range (i.e., 0 to 0.5 m (0 to 19.685 in)) to prevent the robot from detecting its own body as an obstacle. The position and pose of the LiDAR sensor are provided in Table 1, replicating the physical robot.

The model of UR5e robotic arm was adopted and adapted from the Universal Robot's GitHub repository (Husky Manipulation, 2023) within the ROS-Industrial (ROS-Industrial, 2025) framework for integration with the Husky platform. The UR5e is an industrial collaborative robot designed for medium-duty pick-and-place applications with payloads up to 5 kg (11 lb). It features six degree-of-freedom (6-DOF) with a maximum reach of 0.850 m (33.465 in) and is highly configurable with ROS using its dedicated drivers. The UR5e robotic arm was integrated with the Husky robot, with its base positioned and posed at the coordinates given in Table 1. The purpose of integrating the UR5e is to enable cotton picking while autonomously navigating the system in the virtual cotton farm using the Gazebo. Fig. 5 showcases the integrated robotic cotton picker inside the virtual cotton farm from the side view (Fig. 5(a)) and the top view (Fig. 5(b)). The cotton-picking algorithm for this robot is planned to be implemented in our future study.

### 2.5. Cotton-Eye perception system

Cotton-Eye perception system, consisting of three RGB-D cameras, is the key component of this research. It functions as the 'eyes' of the robot, assisting in visual-guided autonomous navigation in the virtual cotton farm, along with other navigation sensors on the robot, e.g., an IMU, a 3D LiDAR, and a vision-RTK. Cotton-Eye also serves as a centralized module to facilitate the robot's navigation (i.e., instance segmentation of the farm scenes) and cotton-picking (i.e., object detection of the cotton bolls) on its traversal path using deep learning-based approach, where YOLOv8 model (Jocher et al., 2023a) was utilized for the segmentation and detection tasks in this study for real-time implementations. The overall workflow of the Cotton-Eye perception system is illustrated in Fig. 6

#### 2.5.1. Cotton boll detection and scene segmentation using YOLOv8

You Only Look Once (YOLO), as one of the most popular deep learning models first introduced in 2016 (Redmon et al., 2016), revolutionized the field of computer vision. Built upon the success of earlier versions of YOLO (Jocher et al., 2023b), YOLOv8 (Jocher et al., 2023a) is an updated and improved version released in 2023. It is considered as one of the most powerful real-time YOLO models, which holds the capability for both multi-class object detection and instance segmentation. Some evidence proved that the YOLOv8 model significantly outperforms the earlier versions of YOLO in terms of accuracy and inference speed (Jocher et al., 2023a; Chen, 2025). The YOLOv8 model can be utilized to perform various computer vision tasks, including object detection, image classification, instance segmentation, pose estimation, and object tracking (Jocher et al., 2023b). Since this research was initiated in early 2023, YOLOv8 was selected as the state-of-the-art model available at the time. Newer versions were not integrated as the study progressed, primarily because the study subsequently prioritized the development and evaluation of autonomous navigation strategies within the simulation framework.

There are five basic YOLOv8 variants with slightly modified architectures and different total number of model parameters, including YOLOv8n (used in this study), YOLOv8s, YOLOv8 m, YOLOv8l, and YOLOv8x (Jocher et al., 2023b). YOLOv8n is a lightweight model capable of performing near real-time detection inference of object detection and instance segmentation with input images. In this study, YOLOv8n was selected to ensure timely decision-making. It is feasible to implement the Cotton-Eye perception system on the cotton-picking robot for real-time navigation control in the virtual cotton farm by leveraging the capability of YOLOv8.

#### 2.5.2. Image acquisition and annotation

The imagery data were collected using a customized ROS Python package script was developed to capture and save images from the primary camera while the robot was traversing the virtual cotton farm in the Gazebo. Two distinct datasets were prepared, where the first dataset consisted of 1000 captured images of the in-field environments (for instance segmentation of farm scenes) and the second dataset consisted of 400 images of the canopy of a single cotton plant (for cotton boll detection) (Fig. 7). For preliminary testing, 80 and 40 images from the first and second datasets, respectively, were randomly selected and manually annotated in Roboflow (Roboflow Inc., Des Moines, IA, USA), as shown in Fig. 7. For the first dataset, the 80 annotated images were further augmented by applying random rotations in the range of ±13°, resulting in a final dataset of 193 images. The first dataset was then split into three subsets, including the train set (173 images), the validation set (16 images), and the test set (4 images), for evaluating the performance of YOLOv8n-seg on segmenting farm environments. For the second dataset, the 40 images of the cotton canopy were captured using the primary camera by maneuvering the cotton-picking robot around the cotton plants and were annotated using Roboflow. The dataset was also divided into three subsets, including the train set (26 images), the validation set (6 images), and the test set (8 images), for evaluating the performance of YOLOv8n on cotton boll detection. Only a subset of images were carefully selected from the larger dataset for model training, because the virtual cotton farm in the Gazebo was relatively uniform in this study.

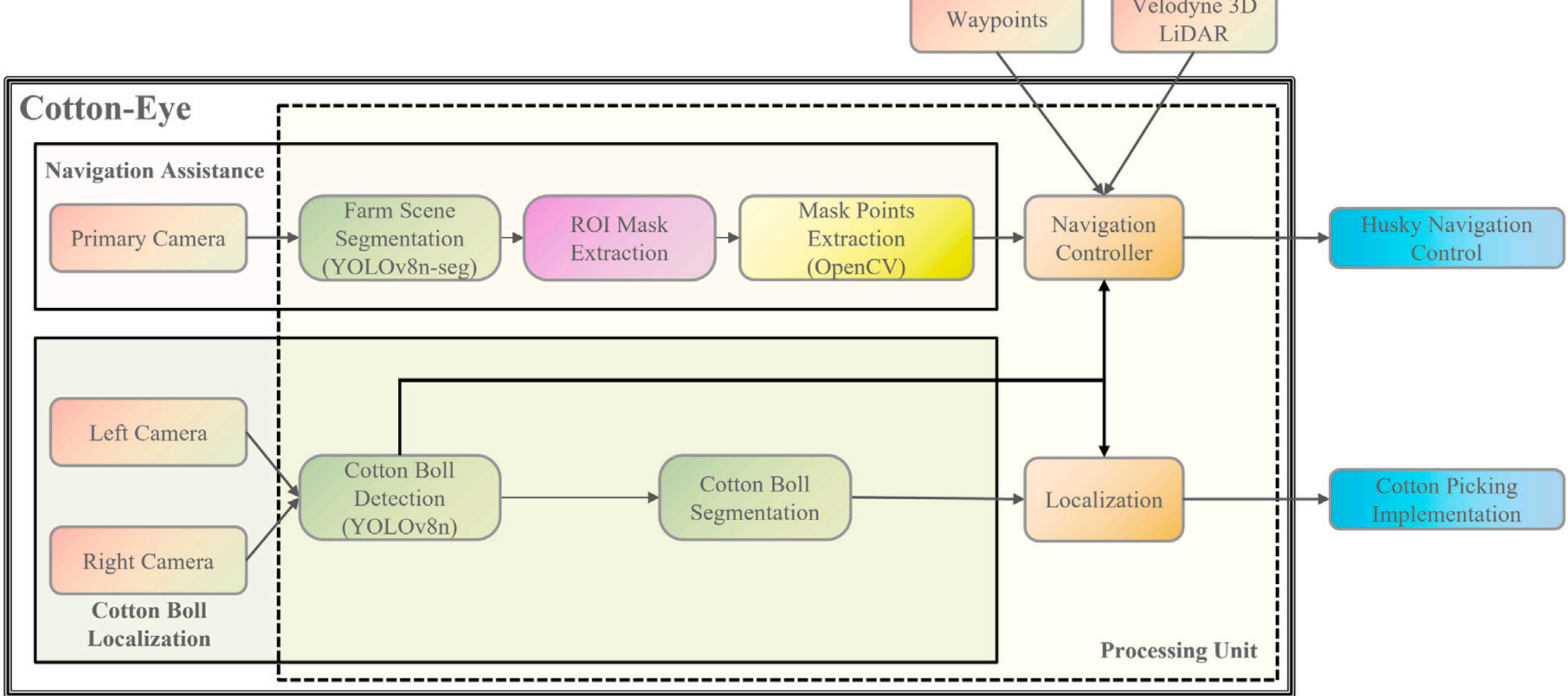

**Fig. 6.** Overall workflow of the Cotton-Eye perception system for the cotton-picking robot (ROI: region of interest).

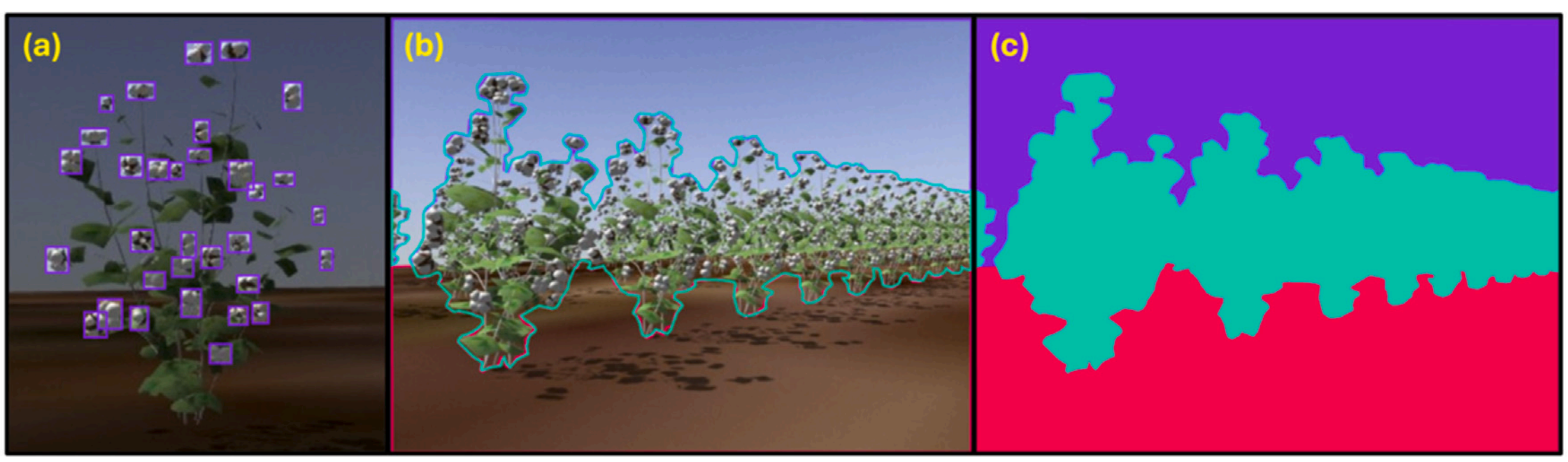

**Fig. 7.** Image dataset annotations: (a) annotation of cotton bolls for detection task, (b) annotation of farm scene for segmentation task, and (c) annotated segmentation masks of 'sky', 'cotton plants', and 'ground' layers.

*2.5.3. Instance segmentation using primary camera*

To facilitate the visual guidance for autonomous navigation of the cotton-picking robot in the virtual cotton farm, the primary camera (24 frames per second (fps)) was employed to continuously generate the ROS image messages, which were converted to OpenCV frames. The converted OpenCV image frames were then inputted into the trained, ready-to-deploy YOLOv8n-seg model with the best trained weights to segment the farm environment. The model was specifically trained to make real-time predictions on three classes, including 'sky', 'ground', and 'cotton plants'. The predicted instance masks from the three classes were extracted to assist in the navigation control of the robot.

*2.5.4. Cotton boll detection using secondary and tertiary cameras*

Cotton boll detection is another crucial task that the Cotton-Eye perception system performs, where the robot needs to halt once the cotton bolls are detected by either the secondary or tertiary camera (24 fps) while the robot traverse the farm. Similarly, these two cameras were used to continuously generating the ROS image messages from both left and right sides of the robot, which were also converted to OpenCV image frames. The image frames were then passed to the trained, ready-to-deploy YOLOv8n model to make real-time predictions on cotton bolls using bounding boxes. The detection results can provide spatial locations of the cotton bolls for robotic picking.

*2.6. Robot navigation control*

The autonomous navigation control of the cotton-picking robot was tested with two approaches: (1) map-based navigation and (2) GPS-based navigation. These two approaches are applied in distinct scenarios, where map-based navigation is often used for small and indoor farms, and GPS-based navigation is used for large farms. To enable efficient navigation control of the robot using these two approaches in this study, the combination of a 3D LiDAR, a vision-RTK, and the Cotton-Eye perception system (with the front-facing primary camera) was utilized. The robot can traverse the virtual cotton farm continuously until its battery level drops to 40%. Upon reaching this level, the robot automatically returns to the 'Home' position for recharging the battery.

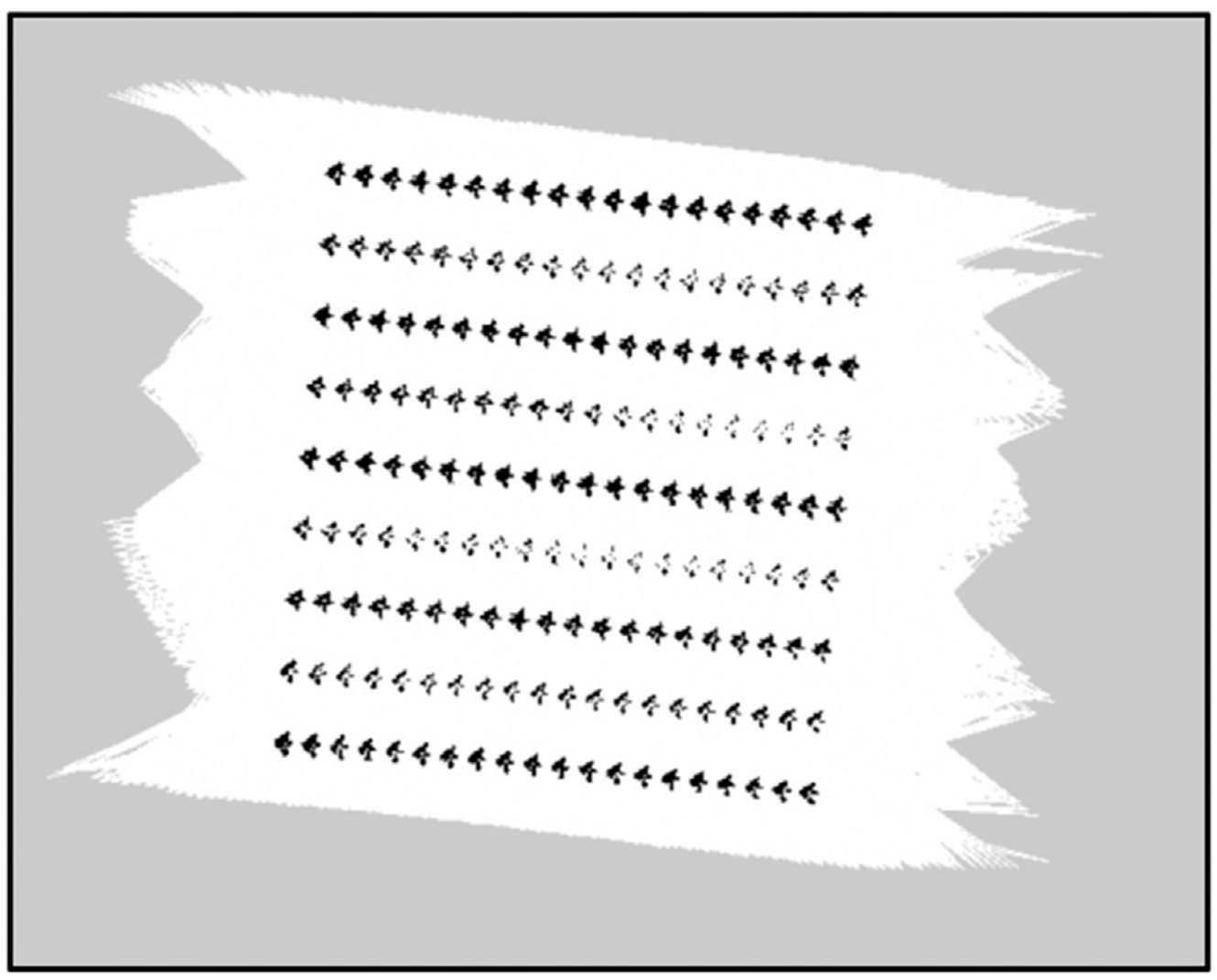

**Fig. 8.** Generated 2D map of the virtual cotton farm using Simultaneous Localization And Mapping (SLAM). Dark-colored pixels represent the cotton plants, while white-colored pixels represent navigable space for the robot.

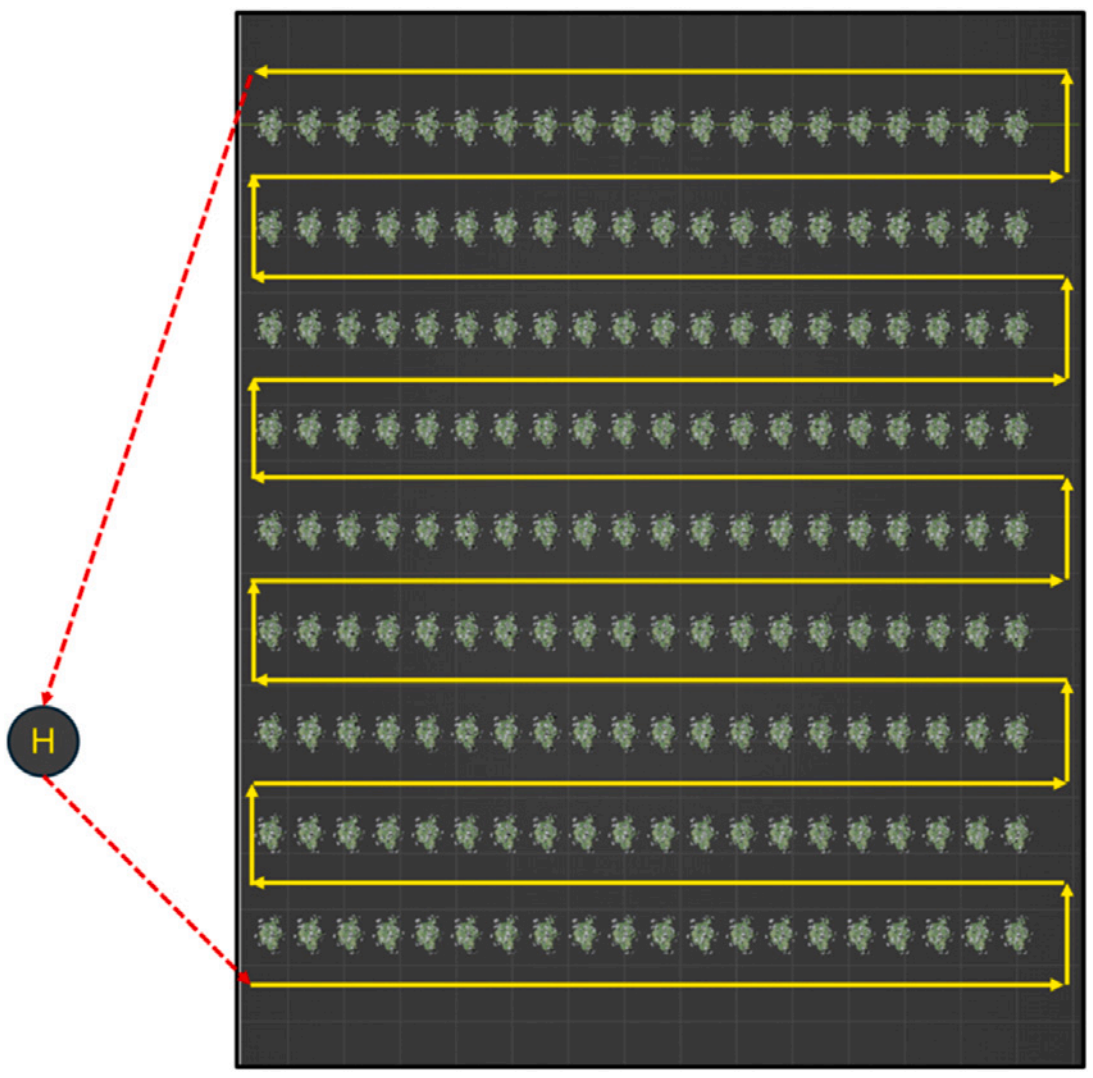

**Fig. 9.** Illustration of the overall navigation path of the cotton-picking robot in the virtual cotton farm. 'H' denotes the 'Home' position.

#### 2.6.1. Map-based navigation

Map-based autonomous navigation control is mainly used for small and indoor farm settings, or when the working area is GPS-denied. This approach requires a pre-generated 2D or 3D map, which is employed to navigate the robot through the farm. In this study, a 3D LiDAR sensor integrated with the Husky robot was used to generate the 2D map of the virtual cotton farm using the Simultaneous Localization And Mapping (SLAM). The mapping process was completed in the Gazebo using the ROS package of '*gmapping*', where the ROS topic of *scan* provided the laser sensor data for constructing the map. The waypoints were then generated from the derived 2D map to control the navigation of the cotton-picking robot in the simulation.

For localization within the generated 2D map, the Adaptive Monte Carlo Localization (AMCL) algorithm was employed using the ROS package *'amcl'* (Husky, 2025), which utilized a particle filter to estimate the robot's position by comparing LiDAR sensor readings with the known map (Table A.4). The motion control of the Husky robot was then executed using the ROS package of *'move_base'*, developed by Clearpath Robotics. It leveraged global and local path planners, specifically *'navfn/NavfnROS'* for global path planning and *'dwa_local_planner/DWAPlannerROS'* for real-time obstacle avoidance (i.e., local path planning). The ROS package '*costmap_2d*' was used to generate global and local costmaps, incorporating LiDAR sensor data for path planning and obstacle detection. Thus, the generated 2D map of the cotton farm (Fig. 8) enabled the autonomous navigation of the cotton-picking robot, assisting in traversing and movement between the selected waypoints in the farm.

#### 2.6.2. GPS-based navigation

For large farm environments, where GPS signals are available, GPS-based navigation is commonly employed using GPS sensors integrated into the system (Griffin, 2009; Mwitta and Rains, 2024). This method utilizes GPS waypoints to guide the robot autonomously across the farm. Initially, an empty map server is initialized in the Gazebo, representing the geographic location of the virtual cotton farm. These coordinates provide the initial reference point for the simulation. The GPS waypoints, provided in latitude and longitude $(L, F)$, are converted into Universal Transverse Mercator (UTM) coordinates $(x, y)$ for more accurate navigation. These UTM coordinates are then mapped onto the initial map server, enabling precise movement within the farm.

Similar to the map-based navigation, the ROS package '*move_base*' was used to guide the robot to its goal positions (i.e., waypoints) through a combination of global and local planners. The navigation system ensured that the robot can move within the cotton rows while progressing toward its next goal position. In this study, the GPS-based navigation involved subscribing to GPS coordinates $(L, F)$ through the *navsat/fix* ROS topic, converting the GPS coordinates into UTM coordinates $(x, y)$ using the 'NavsatConversions' ROS package, and then converting the UTM coordinates to map coordinates using ROS transformations. This provided real-time updates of the robot's location, enabling smooth integration with the *'move_base'* navigation stack and ensuring reliable robot movement across the entire farm.

The Cotton-Eye perception system, with its primary camera, worked together with the GPS waypoints to ensure the robot stays aligned within the crop rows while traversing the farm. Adjustments were made using visual correction algorithms wherever needed. This helped improve navigation efficiency and reduce the risk of crop damage during the autonomous operations.

#### 2.6.3. Overall navigation approach

The predefined map-based coordinates of $(x_i, y_i, \psi_i)$ and a sequence of GPS coordinates $(L_i, F_i)$ were strategically selected to navigate the robot, starting from the entry point of the first cotton row, passing through a sequence of midpoints, and finally reaching the endpoint on the farm. Here, $x$, $y$, and $\psi$ represent the $(x, y)$ coordinates of the map in meters and the yaw angle in radians, whereas $L$ and $F$ represent the latitude and longitude, respectively. The variable $i$ is an integer representing the total number of selected waypoints. The determined traversal path of the robot began at the southeast (SE) corner of the virtual cotton farm for both the map-based and GPS-based approaches, starting from the same location (i.e., 'Home' position with the map coordinates of $(x_h = -6.0, y_h = 1.0, \psi_i = 0)$), progressed from the first row (on the east side) to the last row (on the west side), and concluded at the southwest (SW) corner of the farm at an average traversing speed of 1.8 km per hour. The overall navigation path is illustrated in Fig. 9.

To enable precise visual-guided navigation within the cotton farm, a segmentation-based decision tree approach was developed for each segmented instance. The trained weights of YOLOv8n-seg were implemented for segmentation in the Gazebo virtual cotton farm, resulting in the instance segmentation of 'sky', 'ground', and 'cotton plants'. The predicted segmentation masks were utilized to draw two intersected straight lines, assisting in computing the center of the traversing row using the masks' pixel coordinates of the 'sky' and 'ground', generated by using OpenCV (2025), in each frame. Fig. 10 illustrates the approach of robot pose correction and the process of drawing the two straight lines using the derived masks' coordinates from the YOLOv8n-seg segmentation. Each mask layer was computed to select the maximum $(x_{max}, y_{max})$, mean $(\bar{x}, \bar{y})$, and minimum $(x_{min}, y_{min})$ pixel values, which were used to develop the two intersecting straight lines.

For instance, the 'sky' mask layer's minimum and maximum $y$ pixel coordinates $(y^{sky}_{min}, y^{sky}_{max})$ were selected so that the lower center pixel coordinates and higher center pixel coordinates can be used for drawing a straight line using OpenCV (i.e., two yellow dots in Fig. 10a). Similarly, the lower center pixel coordinates and higher center pixel coordinates of the 'ground' mask $(y^{ground}_{min}, y^{ground}_{max})$ were selected for drawing a straight line using OpenCV (i.e., two red dots in Fig. 10a). These two lines were then extended to the image frame to find their intersection. Fig. 10b showcases the lines and their intersection derived from the 'cotton plants' (the blue line) and 'ground' (the green line) masks. Furthermore, the same process was completed with the 'ground' (the blue line) and 'sky' (the green line) masks to develop the two intersecting lines (Fig. 10c). This study utilized the 'sky' and 'ground' masks for the navigation assistance. The intersection pixel coordinates $(x^{sky-ground}_{int}, y^{sky-ground}_{int})$ were employed to continuously correct the robot's navigation in real-time, guiding it to move forward or turn left or right with a maximum turning angle of 5°. This visual-guided navigation task was executed as the robot traversed from the previous coordinates $(x_i, y_i)$ to the next coordinates $(x_{i+1}, y_{i+1})$ in the virtual cotton farm. The movement decision was made in each image frame to ensure that the robot remained centered within the cotton rows for efficient navigation.

A workstation of Dell (Dell Inc., Round Rock, TX, USA) Precision 7920 Tower (073 A), equipped with an Intel$^{(R)}$ Xeon Gold 6250 CPU (3.90 GHz), 187 GB shared RAM, and a NVIDIA$^{(R)}$ RTX A6000 GPU (49 GB of memory), was used to perform all tasks in this study. The workstation was running the Ubuntu 20.04.5 LTS Linux operating system.

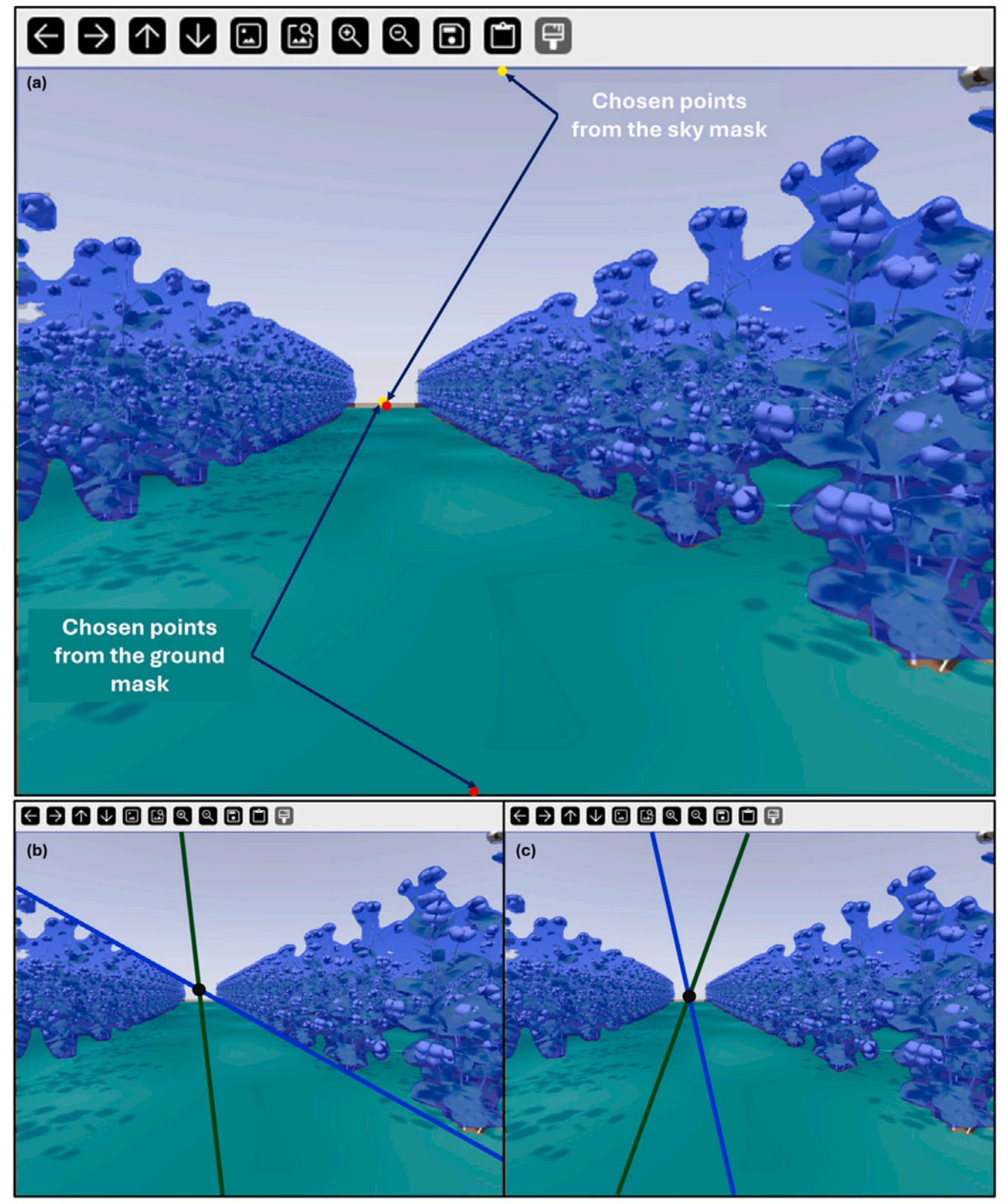

**Fig. 10.** Visual-guided navigation assistance to keep correcting the pose of the cotton-picking robot at the center of the cotton rows in the virtual cotton farm: (a) steps to draw the straight lines derived from the 'sky' and 'ground' segmentation masks, (b) with intersecting lines derived from the 'cotton plants' and 'ground' masks, (c) with intersecting lines derived from the 'sky' and 'ground' masks.

### 2.7. Evaluation metrics

For Cotton-Eye perception system, the performances of YOLOv8n and YOLOv8n-seg models were evaluated using three distinct metrics: $precision$ (Eq. (1)), $recall$ (Eq. (2)), and mean Average Precision $(mAP)$ (Eq. (3)). The $mAP$ refers to the mean of the average precision across correctly predicted classes when the Intersection over Union $(IoU)$ is set to 0.5. The $mAP$ is calculated by averaging the AP values across all classes.

$$Precision = \frac{TP}{TP + FP} \tag{1}$$

$$Recall = \frac{TP}{TP + FN} \tag{2}$$

where TP, FP, and FN denote True Positives, False Positives, and False Negatives, respectively, with regard to the predicted and ground-truth instances.

$$AP = \frac{1}{\sum_{i=1}^{n} r_i} \sum_{i=1}^{n} r_i \left(\frac{\sum_{j=1}^{i} r_j}{i}\right) \tag{3}$$

where $r_i = 1$ if the correct class is detected, or $r_i = 0$ if not.

Additionally, the robot's autonomous navigation trajectory was evaluated against the planned trajectory using the error $(E)$ of the Euclidean distance (Eq. (4)), average error $(AE)$ (Eq. (5)), root mean square error $(RMSE)$ (Eq. (6)), and the completion rate $(CR)$ of the navigation (Eq. (7)). The $E$ quantified the deviation of the robot's actual trajectory from the planned trajectory based on waypoints at each location. The $RMSE$ measured the overall deviation of the actual trajectory from the planned trajectory based on waypoints. Moreover, the $CR$ was used to calculate the percentage of waypoints reached by the robot within an acceptable error margin of 0.25 m for the map-based navigation and (5e−6)° for the GPS-based navigation comparing the planned and actual trajectories, indicating the robot's ability to follow the planned path.

$$E = \sqrt{(x_i - \hat{x}_i)^2 + (y_i - \hat{y}_i)^2} \tag{4}$$

where $(x_i, y_i)$ denotes the planned trajectory of waypoint coordinates, and $(\hat{x}_i, \hat{y}_i)$ denotes the closest actual trajectory of the tracked waypoint coordinates.

$$\text{AE} = \frac{1}{N} \sum_{i=1}^{n} E_i \tag{5}$$

where $N$ denotes the total number of waypoints.

$$RMSE = \sqrt{1/N \sum_{i=1}^{n} (E_i)^2} \tag{6}$$

$$CR = \left(\frac{\sum_{i=1}^{N} \mathbb{1}(E_i \le r)}{N}\right) \times 100 \tag{7}$$

**Table 2**
Training results of the instance segmentation of farm scenes (YOLOv8n-seg) and cotton boll detection (YOLOv8n) tasks in the Gazebo.

| Task | Model | Class | Precision (%) | Recall (%) | mAP (%) (IoU = 0.5) |
|---|---|---|---|---|---|
| Segmentation | YOLOv8n-seg | Sky | 99.0 | 94.1 | 95.0 |
| | | Ground | 99.0 | 94.1 | 95.3 |
| | | Cotton plants | 100.0 | 96.2 | 98.4 |
| | | All | 99.3 | 94.8 | 96.2 |
| Detection | YOLOv8-n | Cotton boll | 92.9 | 85.6 | 92.6 |

**Table 3**
Testing results of the instance segmentation of farm scenes (YOLOv8n-seg) and cotton boll detection (YOLOv8n) tasks in the Gazebo.

| Task | Model | Class | Precision (%) | Recall (%) | mAP (%) (IoU = 0.5) |
|---|---|---|---|---|---|
| Segmentation | YOLOv8n-seg | Sky | 99.1 | 66.7 | 66.9 |
| | | Ground | 99.7 | 100 | 99.5 |
| | | Cotton plants | 82.1 | 100 | 86.2 |
| | | All | 93.0 | 88.9 | 85.2 |
| Detection | YOLOv8-n | Cotton boll | 94.8 | 87.2 | 92.7 |

where $\mathbb{1}(E_i \leq r)$ is an indicator function that returns 1 if the waypoint error $E_i$ is within the threshold $r$, or 0 if not.

## 3. Results

### 3.1. YOLOv8 model training and validation

With the Cotton-Eye perception system, the YOLOv8n-seg model achieved desired instance segmentation outcomes in the virtual cotton farm. During the training and validation process, the overall $precision$, $recall$, and $mAP$ were 99.3%, 94.8%, and 96.2%, respectively, for all classes, including the 'sky', 'ground', and 'cotton plants'. The specific class-wise training results are given in Table 2. Additionally, the YOLOv8n model demonstrated exceptional performance in detecting cotton bolls as part of the Cotton-Eye perception system's functionality. The $precision$, $recall$, and $mAP$ for cotton boll detection were 92.9%, 85.6%, and 92.6%, respectively.

### 3.2. YOLOv8 model testing

Similarly, the YOLOv8n-seg model achieved good testing results for instance segmentation. The overall $precision$, $recall$, and $mAP$ were 93.0%, 88.9%, and 85.2%, respectively, for all three classes. More specifically, the class-wise testing results are presented in Table 3. Furthermore, the $precision$, $recall$, and $mAP$ testing metrics for cotton boll detection were 94.8%, 87.2%, and 92.7%, respectively.

### 3.3. Autonomous navigation in virtual cotton farm

The cotton-picking robot was successfully launched at the 'Home' position $(x_h, y_h)$ in the Gazebo virtual cotton farm. The UR5e arm's driver was launched simultaneously, marking the success of the very first integration of this robotic arm and the Husky robotic mobile platform in the agricultural engineering community, which can be controlled to perform cotton-picking tasks in the Gazebo. For the map-based approach, waypoints — defined by $(x, y)$ coordinates in meters based on the map frame, along with orientation in radians — were set to guide the robot into the cotton farm from the southeast (SE) corner for autonomous navigation. Similarly, for the GPS-based approach, GPS waypoints — i.e., coordinates containing latitude and longitude — were initialized to achieve the same objective. Collectively, with the support of the 3D LiDAR, Cotton-Eye perception system, and map-based or GPS-based navigation control method, the robot successfully and efficiently navigated itself to traverse the entire virtual cotton farm (e.g., entering, moving, exiting, turning, re-entering, moving, re-exiting, etc.) as showcased in Fig. 11.

Overall, implementing either map-based or GPS-based navigation approach resulted in successful and smooth autonomous navigation of the cotton-picking robot in the virtual cotton farm, with its visual-guided navigation assistance continuously correcting the robot's pose and aligning the robot at the center of the cotton rows. First, the primary camera of the Cotton-Eye perception system performed satisfactorily using the trained YOLOv8n-seg model for instance segmentation of the farm environment in the Gazebo to visually guide the robot. The segmentation results were showcased in Fig. 12, where a clear division of the 'sky', 'ground', and 'cotton plants' was observed. Furthermore, the secondary and tertiary cameras were also highly effective in detecting cotton bolls, using the trained YOLOv8n model, while the robot was traversing the virtual cotton farm. The detection results were showcased in Fig. 13, demonstrating the capability of the Cotton-Eye perception system for identifying the target cotton bolls in real-time.

The traversal path of the cotton picker was monitored with the planned waypoints to monitor the trajectory of the cotton picker. Fig. 14 shows the trajectory of the cotton picker using GPS-based (Fig. 14a) and map-based (Fig. 14b) autonomous navigation. Expected trajectory is given in red color and the planned trajectory is given in blue color.

Finally, the robot took an average of 31 min and 7 s to completely navigate through the entire cotton farm using the map-based navigation. The $AE$ was 0.137 m for all planned waypoints comparing against the nearest actual traversed waypoints, while the $RMSE$ was 0.156 m for the overall navigation trajectory. Additionally, a 96.7% of $CR$ was recorded with a threshold value of 0.25 m for the nearest tracked waypoint. Meanwhile, the robot took an average of 20 min and 42 s to complete the navigation using the GPS-based approach. The $AE$, $RMSE$, and $CR$ with a threshold value of (5e−6)° were 0 m, 0 m, and 100%, respectively.

## 4. Discussion

### 4.1. Cotton boll detection and scene segmentation

Overall, the visual-guided navigation performance of our cotton-picking robot in the Gazebo virtual cotton farm was notably robust and reliable despite the Cotton-Eye perception system being trained with a limited number of images in the dataset, where a total of 173 and 16 images were used in the training and testing sets for segmenting the farm environment. The segmentation and detection results, reflected in the desired testing $mAP$ of 85.2% and 92.7% (Table 3), demonstrated that the perception system could accurately identify the cotton farm features (e.g., 'sky', 'ground', 'cotton plants', and cotton bolls) for effective navigation assistance. Moreover, the adoption of YOLOv8n model proved beneficial by offering comparatively faster inference speed with less number of model parameters (i.e., 3.2 million Jocher et al., 2023b). With an average segmentation inference speed of approximately 126.8 ms per image, the robot was capable of making near real-time movement decisions, thereby enhancing overall operational responsiveness in the simulation.

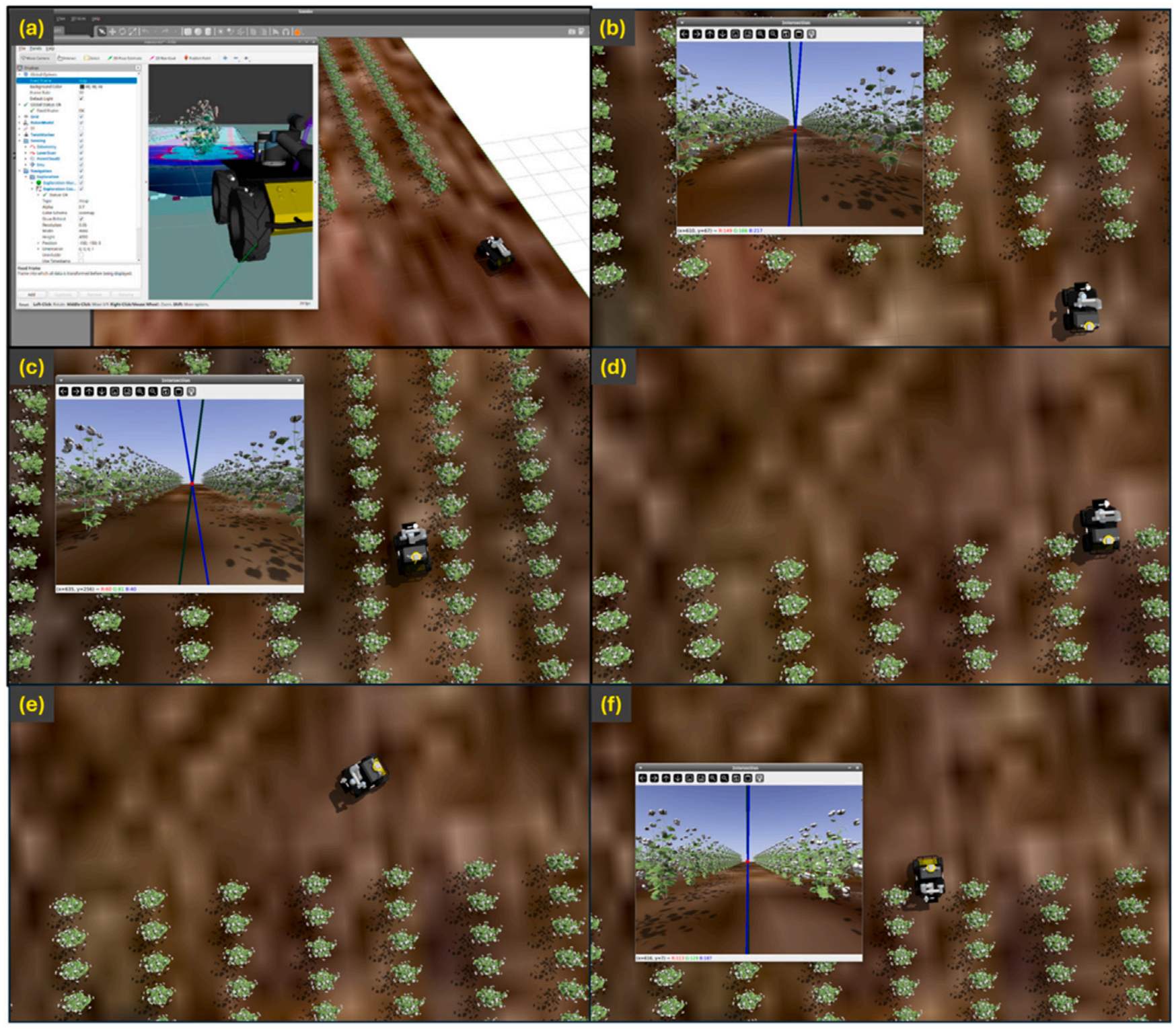

**Fig. 11.** Illustrations of the cotton-picking robot autonomously navigating itself in the virtual cotton farm in the Gazebo: (a) starting the navigation, (b) entering a row, (c) moving through the row, (d) exiting the row, (e) making a turn to the next row, and (f) re-entering the next row. The video demonstrations are available at GitHub (https://github.com/imtheva/CottonSim).

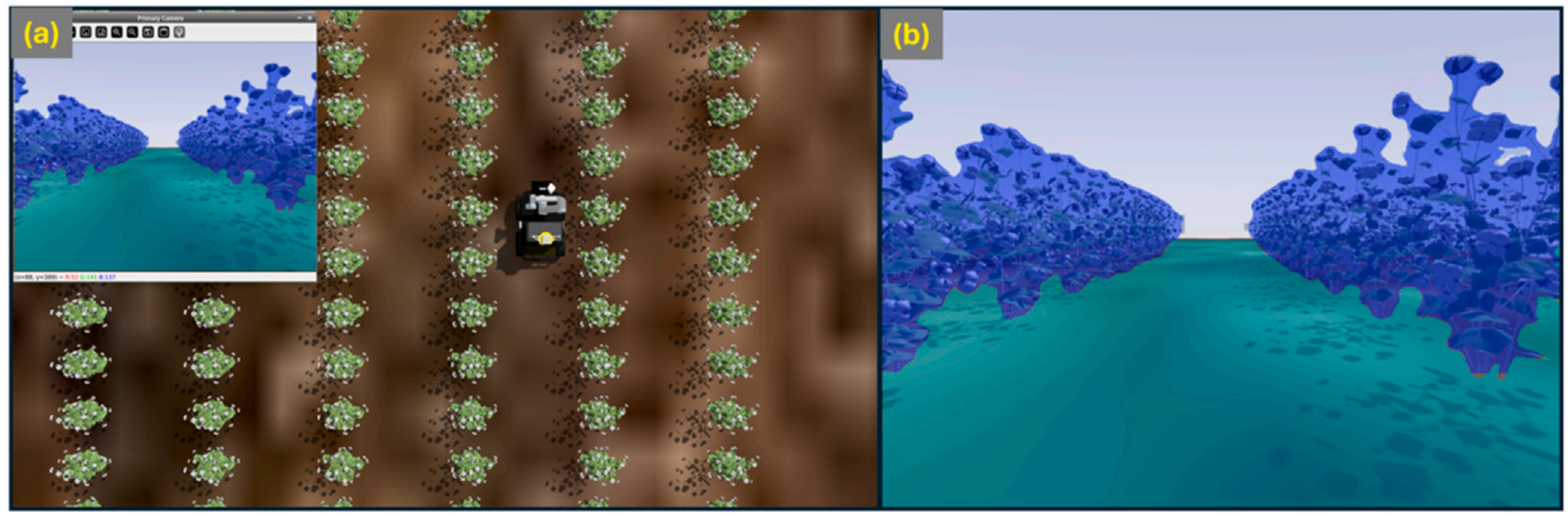

**Fig. 12.** Deployment of the (a) Cotton-Eye perception system on the cotton-picking robot for (b) instance segmentation of the farm environment using the YOLOv8n-seg model with the front-facing primary camera in Gazebo.

Additionally, cotton boll detection and scene segmentation were relatively straightforward in the simulation environments, as several critical physical factors such as terrain variations, frictional forces, and wind disturbances remained unchanged. The high accuracies of detection and segmentation observed in this setting were largely attributed to the simplicity of the virtual environment. However, real-farm conditions would introduce additional complexities, necessitating further calibration to mitigate the impact of those physical factors on perception and navigation. Under ideal conditions, cotton boll detection is easier when bolls are fully exposed, illuminated by uniform lighting, and contrasted distinctly against the background.

However, transitioning from simulation to real-world deployment may introduce substantial challenges. Field environments are dynamic and unpredictable, making object detection significantly more difficult. Factors such as uneven terrain, inconsistent illumination, shadows, and occlusions from overlapping or dense vegetation can severely impact visibility and detection accuracy. In addition, external environmental conditions, such as changing weather, dust accumulation on sensor surfaces, and fluctuating light intensities, can further degrade the performance of perception systems.

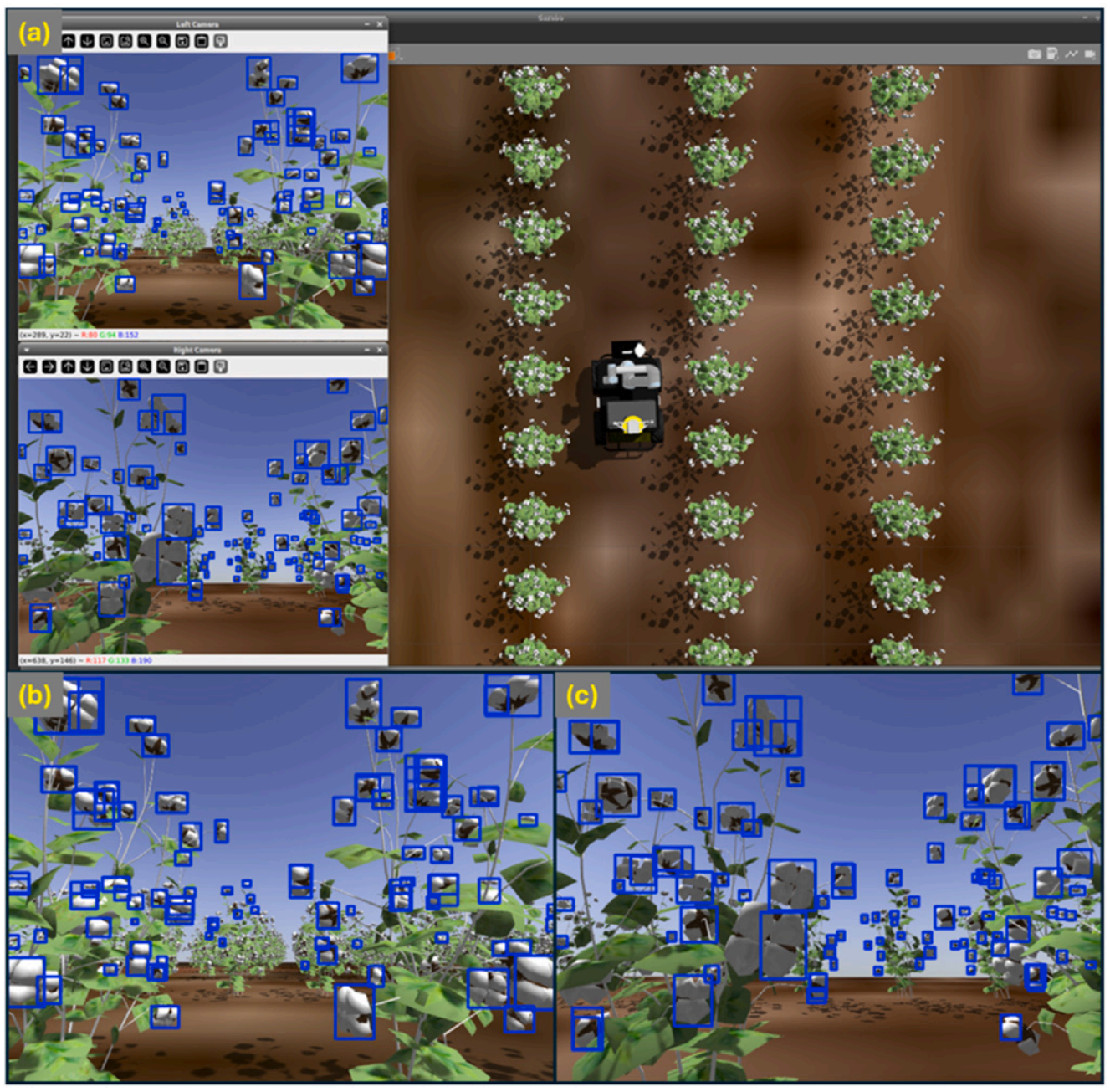

**Fig. 13.** Deployment of the (a) Cotton-Eye perception system on the cotton-picking robot for (b) cotton boll detection using the YOLOv8n model with the (a) left-facing secondary camera and (c) right-facing tertiary camera in Gazebo.

Given that these challenges are absent in simulation, real-world deployments require more robust and adaptive perception strategies to ensure reliable system performance. Addressing these challenges may involve the use of illumination-invariant deep learning models, advanced background subtraction techniques, and domain adaptation methods to enable consistent perception under varying agricultural conditions. These enhancements will be essential for bridging the gap between simulation-based development and practical field implementation.

In addition to perception, planting configurations in real-world cotton farms present their own challenges. The current system was initially tested in a simulated environment with well-spaced rows, which may not fully reflect dense or irregular planting conditions found in practice. Our robot is designed to operate in fields with slightly wider row spacing to accommodate safe navigation and precise arm articulation. However, the navigation and picking modules are modularized and can be adapted to smaller robots optimized for narrower planting configurations. In future implementations, adjustments to row spacing and planting layout may be considered to maximize operational efficiency, or alternative robot designs, such as over-the-row (OTR) system, may be deployed to suit other existing field configurations.

### 4.2. Robot navigation strategies

The navigation strategy for the cotton-picking robot was implemented using both map-based and GPS-based methods, each commencing from a common 'Home' position in the virtual cotton farm. In the map-based approach, the robot autonomously traversed the mapped area through the selected waypoints (Fig. 14b), while the GPS-based approach followed pre-planned waypoints across the entire farm (Fig. 14a). Both navigation approaches allowed the robot to navigate through the cotton field effectively. However, GPS-based navigation clearly showed more prominent results in multiple test runs compared to map-based navigation, despite the presence of GPS drifting issues.

The cotton-picking robot completed the navigation in the virtual cotton farm 10 min and 25 s faster using GPS-based approach compared to map-based approach. Although both approaches utilized visual-guided navigation assistance that employed a segmentation-based strategy to continue aligning the robot at the center of the cotton rows, map-based navigation took much longer and involved several deviations and unexpected turns along the planned trajectory waypoints (Fig. 14b). In contrast, the robot smoothly traversed the cotton field using GPS-based navigation (Fig. 14a).

### 4.3. Navigation performance evaluation

The evaluation of navigation accuracy highlighted the differences between the two autonomous navigation approaches adopted by the cotton-picking robot. The map-based navigation showed an $AE$ of 0.137 m (Eq. (5)) for all planned waypoints versus their nearest actual traversed counterparts, with an $RMSE$ of 0.156 m (Eq. (6)) along the entire navigation trajectory. Despite these deviations, its $CR$ was 96.7% (Eq. (7)), showcasing a high level of adherence to the planned trajectory with the 0.25 m threshold. Comparatively, GPS-based navigation demonstrated a greater accuracy, achieving an $AE$ of 0 m, an $RMSE$ of 0 m, and a 100% $CR$ with the threshold of (5e−6)°. These results indicated that GPS-based approach ensured near-perfect adherence to the planned trajectory, eliminating positional deviations that were observed in the map-based approach.

The observed differences can be attributed to the inherent limitations of map-based navigation, which relied on pre-defined trajectory planning and environmental perception for localization. Factors such as sensor noises, environmental variations, and trajectory realignment contributed to slight deviations and increased the total navigation time in the cotton farm. Conversely, GPS-based navigation benefited from real-time positional updates with high accuracy, allowing the robot to maintain an optimal trajectory with minimal corrections. These findings suggested that GPS-based navigation, when available, would be more efficient for autonomous cotton farming, ensuring faster and more precise movement through the entire field.

Although GPS drift was observed in this study, the navigation results were not negatively affected. However, GPS readings in the real-world would be less accurate compared to those in the Gazebo simulation. In practice, real-farm GPS data — including RTK-GPS — can suffer from minor deviations due to poor correction signals or the failure to receive corrected coordinates.

Moreover, using 3D LiDAR offered significant advantages over 2D LiDAR. While the 3D LiDAR demonstrated a better performance in SLAM using map-based navigation, its capability of obstacle avoidance was implemented but not evaluated in this study. The SLAM map generated using the 3D LiDAR *'scan'* topic (Table A.4) exhibited a much higher resolution (Fig. 8) compared to the 2D LiDAR-based SLAM approach (not demonstrated in this study). Additionally, 3D LiDAR scanning improved map-based navigation with AMCL (Robot Navigation Control section), ultimately enhancing overall navigation control. The map-based approach would be particularly suitable for indoor environments, greenhouses, small-scale farms, and other GPS-denied environments, where 3D LiDAR-based navigation can enable precise localization.

In agriculture, numerous simulation-based robotic approaches have been developed for tasks such as navigation, mapping, and crop monitoring. Only a limited number of studies have specifically focused on cotton farms. Although several cotton-picking robots have been introduced in recent years (Wang et al., 2008; Fue et al., 2019; Singh et al., 2024; Gharakhani et al., 2024), these efforts primarily emphasized hardware design and field testing. Notably, none of them incorporated a fully autonomous robotic simulation framework tailored for cotton-picking applications. In this context, our study presents one of the first simulation-based implementations designed to evaluate and refine autonomous navigation strategies for cotton-picking robots. This preliminary work aims to provide a foundational testbed that can help identify performance bottlenecks and key areas for improvement prior to real-world deployment.

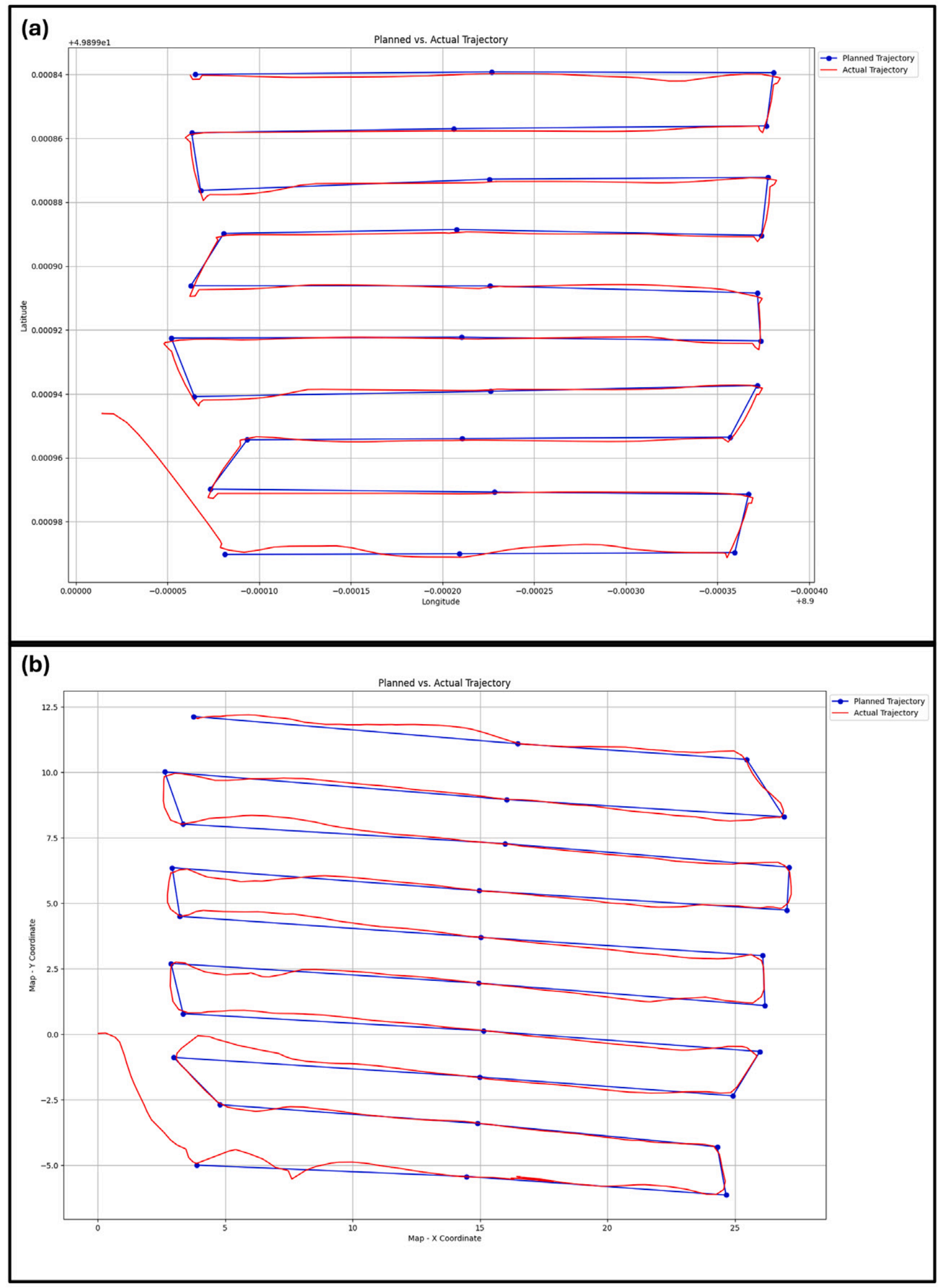

**Fig. 14.** The traversal trajectories of the cotton-picking robot in the virtual cotton farm using the (a) GPS-based and (b) map-based navigation. The video demonstrations are available at GitHub (https://github.com/imtheva/CottonSim).

While these findings highlight the potential of the proposed navigation strategies in a controlled simulation environment, several limitations regarding system performance and simulation conditions must be acknowledged. Although the development of the CottonSim framework was successful, the robot's navigation performance was notably affected by odometry and GPS drift. This included both inherent sensor drift and artificially introduced noise to simulate real-world conditions. These issues complicated the definition of waypoints in both map and GPS coordinate frames, which was an essential aspect for autonomous navigation and perception.

To maintain effective alignment with the Cotton-Eye perception system, waypoint positions had to be manually estimated by driving the robot throughout the virtual cotton farm. Despite this effort, completely mitigating drift remained a challenge, even when using data from the simulated Husky robot's wheel encoders, IMU, and GPS sensors.

Additionally, the simulation framework imposed significant computational demands. Unlike prior studies that employed simple bounding boxes to represent field obstacles, this work used detailed mesh models of cotton plants and generated obstacle boundaries directly around those meshes. To manage computational load, the plant arrangement in the virtual field was kept uniformly spaced, which reduced realism but allowed smoother simulation performance. Altering the position and orientation of individual cotton plant meshes was avoided to prevent further increases in simulation complexity.

## 5. Conclusions

In this study, CottonSim, a Gazebo-based simulation framework for selective cotton harvesting, was developed to address the lack of domain-specific robot simulation platforms. The primary contributions include the design of a lightweight autonomous cotton-picking robot equipped with a UR5e arm, 3D LiDAR, RGB-D cameras, GPS, and IMU, as well as the construction of a realistic virtual cotton farm in ROS-Gazebo. Two autonomous navigation approaches, i.e., map-based and GPS-based, were implemented to evaluate their applicability in cotton farm environments. The key conclusions are as follows:

- A modular, lightweight cotton-picking robot and a virtual cotton farm were successfully developed within the ROS Gazebo simulator. The robot was equipped with a UR5e manipulator, 3D LiDAR, GPS, IMU, and RGB-D cameras, forming a comprehensive hardware stack for autonomous operation.
- The vision-based Cotton-Eye perception system was integrated into the robot to support real-time visual processing. Its navigation assistance module, driven by the YOLOv8n-seg model, achieved a $precision$ of 93.0%, $recall$ of 88.9%, and $mAP$ of 85.2%, enabling robust segmentation of the farm environment. The cotton boll localization module using YOLOv8n achieved a $precision$ of 94.8%, $recall$ of 87.2%, and $mAP$ of 92.7%, demonstrating high accuracy in detecting seed cotton bolls.
- The GPS-based navigation approach significantly outperformed the map-based method, completing the total map path 10 min and 25 s faster and demonstrating improved adherence to the planned trajectory.
- Quantitatively, the GPS-based system achieved perfect navigation metrics ($AE = 0$ m, $RMSE = 0$ m, $CR = 100\%$), underscoring its superior accuracy. When paired with the Cotton-Eye perception system, the GPS-based approach proved highly effective for potential deployment in real-world cotton harvesting scenarios.

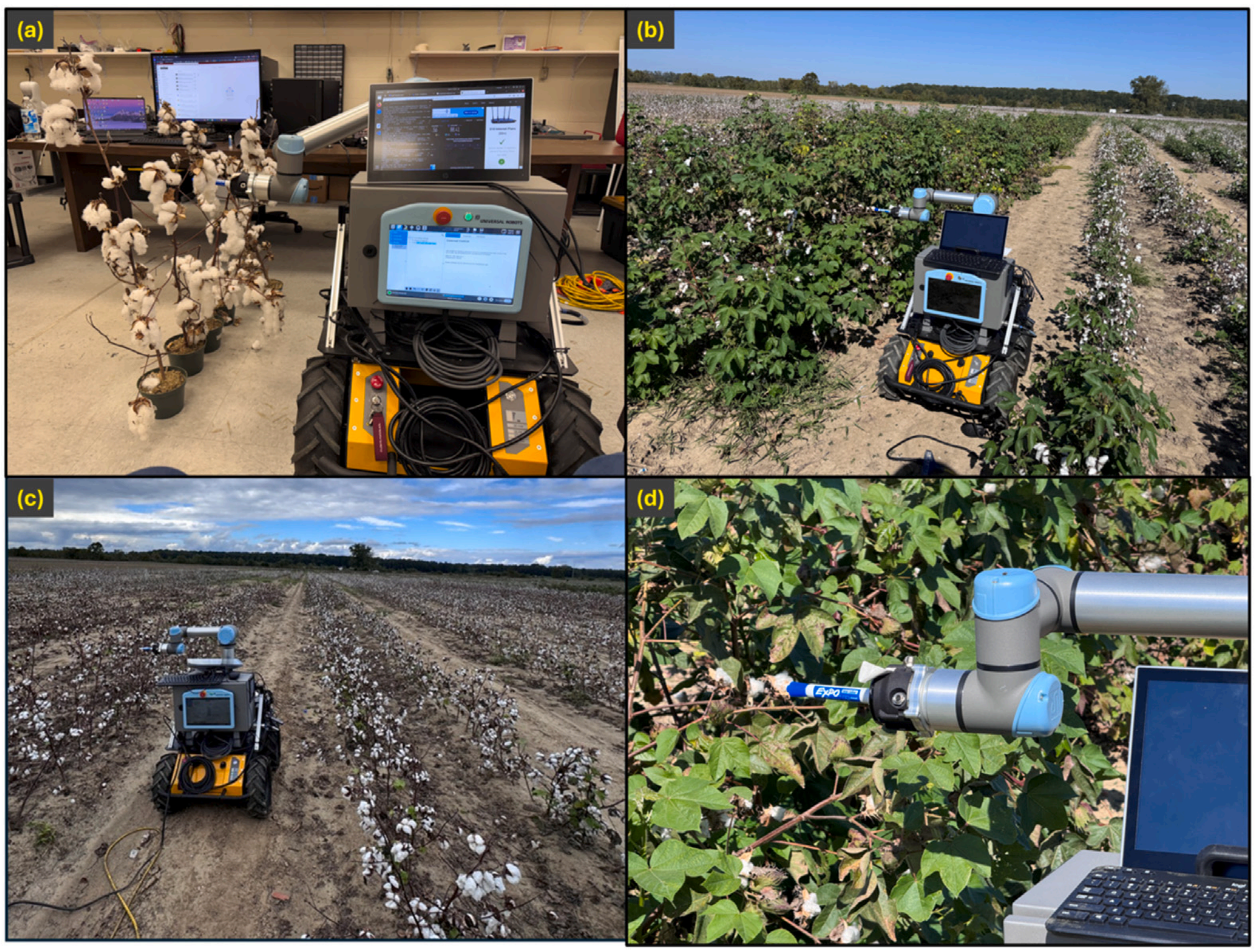

**Fig. 15.** Development and evaluation of the physical cotton-picking robot in the (a) lab setting and cotton farm (b) before and (c) after defoliation. (d) A close-up look of the robot in the field.

To conclude, the objectives of this study were fully realized through the development, integration, and evaluation of a simulation-based, vision-guided robotic cotton-picking system. CottonSim provides a versatile testbed for evaluating autonomous navigation and perception strategies, offering a cost-effective and risk-free environment for iterative development. The promising results of the GPS-based method, enhanced by visual segmentation and detection, suggest a viable path toward real-world deployment.

## 6. Future work

Future work will focus on refining various navigation algorithms to enhance the robot's turning efficiency, enabling more effective and precise movement. Additionally, we will integrate sensor fusion with 3D SLAM to improve localization within the farm environment. Incorporating visual odometry calculations using a camera will be advantageous in addressing and minimizing odometry drift.

Furthermore, integrating the cotton-picking mechanism into 'CottonSim' is essential, as the study aims to achieve both autonomous navigation and cotton picking. While the current study employed the YOLOv8n models for real-time cotton boll detection and farm scene segmentation, future research will investigate newer YOLO versions to assess their potential in improving perception performance under dynamic field conditions. Additionally, complementary segmentation models such as the Segment Anything Model (SAM) (Kirillov et al., 2023) will be evaluated to enable precise segmentation of seed cotton fibers and facilitate accurate end-effector alignment during picking. The simulation algorithms from 'CottonSim' will also be implemented in the corresponding physical system (Fig. 15), which is currently being developed by the research team. As a preliminary validation step, the robot was brought into actual cotton fields before and after the defoliation stage (Fig. 15b–c) to evaluate its picking capabilities in a real-world setting. However, full autonomous navigation and perception testing remain part of our future work.

**Table A.4**
Summary of ROS topics and namespaces developed in this study.

| Component/Functionality | Namespaces | ROS topics |
|---|---|---|
| Primary camera | */realsense_p* | */color/image_raw*; */depth/color/points* |
| Secondary camera | */realsense_secondary* | */color/image_raw*; */depth/color/points* |
| Tertiary camera | */realsense_tertiary* | */color/image_raw*; */depth/color/points* |
| IMU | */imu* | */data* |
| GPS | */navsat* | */fix* |
| LiDAR | */vlp* | */scan* |
| | */points* | – |
| | */scan* | – |
| UR5e arm | */joint_states* | – |
| Robot control | */amcl* | */amcl_pose* |
| Robot navigation | */move_base* | */status* |
| | */move_base_simple* | */goal* |
| Odometry | */odometry* | */filtered* |
| PassThrough filter | */passthrough* | */output* |
| Instance segmentation | */mask* | */intersection/centerskyground* |
| Navigation assistance | */husky* | */cv/coords* |

## CRediT authorship contribution statement

**Thevathayarajh Thayananthan:** Writing – original draft, Visualization, Validation, Software, Methodology, Investigation, Formal analysis, Data curation, Conceptualization. **Xin Zhang:** Writing – review & editing, Supervision, Resources, Project administration, Methodology, Investigation, Funding acquisition, Conceptualization. **Yanbo Huang:** Writing – review & editing, Resources. **Jingdao Chen:** Writing – review & editing, Supervision. **Nuwan K. Wijewardane:** Writing – review & editing, Supervision. **Vitor S. Martins:** Writing – review & editing, Supervision. **Gary D. Chesser:** Writing – review & editing, Supervision. **Christopher T. Goodin:** Writing – review & editing, Supervision.

## Declaration of Generative AI and AI-assisted technologies in the writing process

During the preparation of this work the authors used ChatGPT in order to improve the readability and language of the manuscript. After using this tool/service, the authors reviewed and edited the content as needed and take full responsibility for the content of the published article.

## Declaration of competing interest

The authors declare that they have no known competing financial interests or personal relationships that could have appeared to influence the work reported in this paper.

## Acknowledgments

This work is supported by the Cotton Incorporated (Grant No. 23-889 and 25-582). The authors would like to thank the support from the Mississippi Agricultural and Forestry Experiment Station (MAFES) and the U.S. Department of Agriculture (USDA) Hatch Multi-State project under accession number MIS-402060. Any opinions, findings, conclusions, or recommendations expressed in this publication are those of the authors and should not be construed to represent any views of the funding agency and/or the institutions that the authors are affiliated with.

## Appendix A. ROS topics and namespaces

See Table A.4

## Appendix B. Supplementary material

All models and use instructions are available here on GitHub: https://github.com/imtheva/CottonSim All datasets are available here: https://github.com/imtheva/CottonSim/tree/main/datasets All videos are available here: https://github.com/imtheva/CottonSim/tree/main/Videos.

## Data availability

All datasets, models, and videos are available at https://github.com/imtheva/CottonSim, with use instructions provided in the repository documentation.